\documentclass[sigconf, screen]{acmart}
\usepackage{cases, algorithm, algcompatible, bm, amsmath,multirow}
\usepackage{threeparttable}
\usepackage{pifont, hyperref}

\AtBeginDocument{%
  }

\copyrightyear{2024} 
\acmYear{2024} 
\setcopyright{acmlicensed}\acmConference[ICCAD '24]{IEEE/ACM International Conference on Computer-Aided Design}{October 27--31, 2024}{New York, NY, USA}
\acmBooktitle{IEEE/ACM International Conference on Computer-Aided Design (ICCAD '24), October 27--31, 2024, New York, NY, USA}
\acmDOI{10.1145/3676536.3676796}
\acmISBN{979-8-4007-1077-3/24/10}

\begin{document}


\title[Fast and Efficient 2-bit LLM Inference on GPU: 2/4/16-bit in a Weight Matrix with Asynchronous Dequantization]{Fast and Efficient 2-bit LLM Inference on GPU: \\ 2/4/16-bit in a Weight Matrix with Asynchronous Dequantization}

\author{Jinhao Li$^{1*}$, Jiaming Xu$^{13*}$, Shiyao Li$^{23}$, Shan Huang$^1$, Jun Liu$^1$, Yaoxiu Lian$^1$, Guohao Dai$^{13\dag}$}

\affiliation{%
  \country{$^1$Shanghai Jiao Tong University, $^2$Tsinghua University, $^3$Infinigence-AI, $^*$Equal Contributions} 
}

\affiliation{%
  \country{ $^\dag$Corresponding author: daiguohao@sjtu.edu.cn}
}

\renewcommand{\shortauthors}{Jinhao Li et al.}


\begin{abstract}
  Large language models (LLMs) have demonstrated impressive abilities in various domains while the inference cost is expensive.
  Many previous studies exploit quantization methods to reduce LLM inference cost by reducing latency and memory consumption.
  Applying 2-bit single-precision weight quantization brings $>$3\% accuracy loss, so the state-of-the-art methods use mixed-precision methods for LLMs (\textit{e.g.} Llama2-7b, etc.) to improve the accuracy.
  However, challenges still exist:
  \textbf{(1) Uneven distribution in weight matrix.}
  Weights are quantized by groups, while some groups contain weights with large range. Previous methods apply inter-weight mixed-precision quantization and neglect the range difference inside each weight matrix, resulting in $>$2.7\% accuracy loss (\textit{e.g.} LLM-MQ and APTQ).
  \textbf{(2) Large speed degradation by adding sparse outliers.}
  Reserving sparse outliers improves accuracy but slows down the speed affected by the outlier ratio (\textit{e.g.} 1.5\% outliers resulting in $>$30\% speed degradation in SpQR).
  \textbf{(3) Time-consuming dequantization operations on GPUs.}
  Mainstream methods require a dequantization operation to perform computation on the quantized weights, and the 2-order dequantization operation is applied because scales of groups are also quantized.
  These dequantization operations lead to $>$50\% execution time.
  
  To tackle these challenges and enable fast and efficient LLM inference on GPUs, we propose the following techniques in this paper.
  \textbf{(1) Intra-weight mixed-precision quantization.}
  We only quantize a small fraction of groups with higher sensitivity (larger Hessian value and range variation) using 4-bit.
  Meanwhile, we also take the memory alignment into consideration on GPUs.
  \textbf{(2) Exclusive 2-bit sparse outlier with minimum speed degradation.}
  We only reserve a small fraction of large weights in 2-bit groups as sparse outliers using 16-bit, which leads to a lower average bit increment and speed degradation.
  \textbf{(3) Asynchronous dequantization.}
  We point out that calculating the scales of each group in 2-order dequantization is independent of the loading weights of each group in 1-order dequantization.
  Thus, we design the asynchronous dequantization on GPUs.
  We conduct extensive experiments on different model families (\textit{e.g.} Llama3, etc.) and model sizes.
  We achieve 2.91-bit for each weight considering all scales/zeros for different models with negligible loss.
As a result, with our 2/4/16 mixed-precision quantization for each weight matrix and asynchronous dequantization during inference, our design achieves an end-to-end speedup for Llama2-7b is 1.74$\times$ over the original model, and we reduce both runtime cost and total cost by up to 2.53$\times$ and 2.29$\times$ with less GPU requirements.
\end{abstract}

\maketitle

\begin{figure}[!t]
  \centering  
  \vspace{0pt}
  \includegraphics[width=0.48\textwidth]{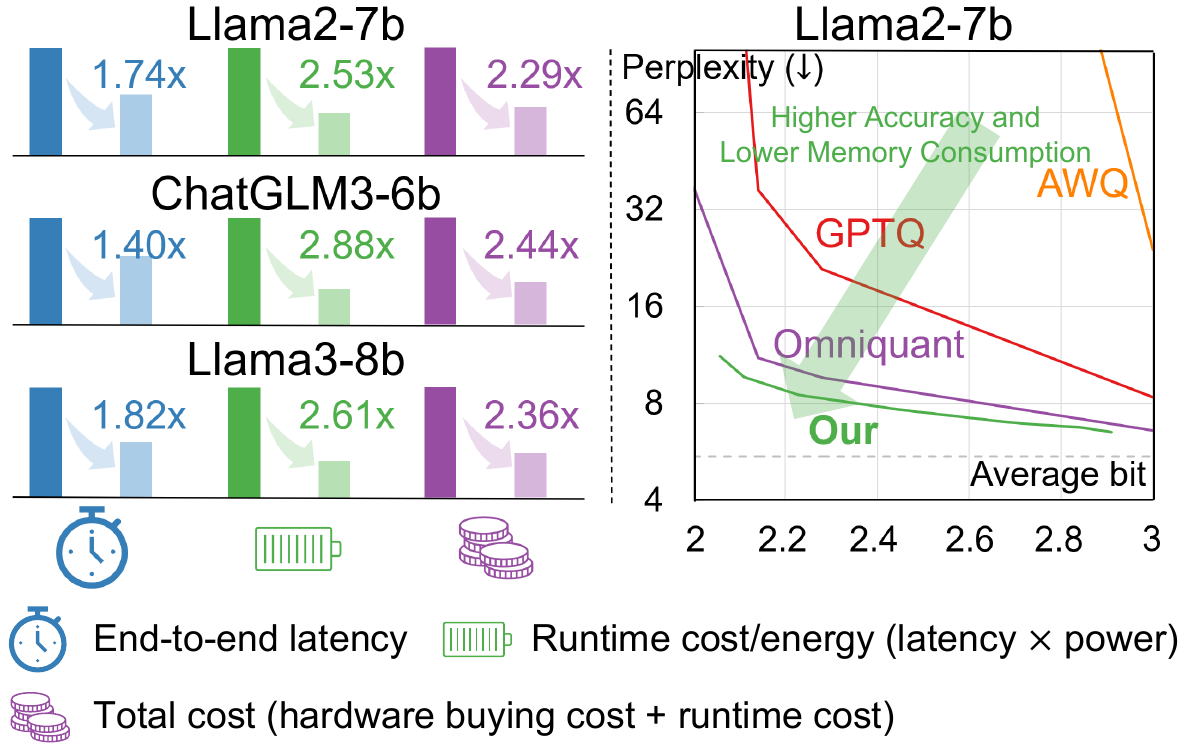}
  \vspace{-20pt}
  \caption{Summary of this paper compared with state-of-the-art quantization designs on mainstream large language models. Detailed data can be found in Table II and Section~\ref{sec:setup}.}
  \vspace{-20pt}
  \label{fig:cost_comparison}
\end{figure}

\section{Introduction}
Large language models (LLMs) demonstrate remarkable capabilities in various domains, excelling in tasks like natural language understanding and generation~\cite{nlu,nlg,min2023recent,lund2023chatgpt,liu2024your}.
However, their computational expense during inference is noteworthy.
For instance, OpenAI reveals that the GPT-4 Turbo incurs a cost of millions of dollars each day~\cite{openai_cost}.
This substantial cost underscores the economic considerations associated with deploying and utilizing such advanced language models~\cite{gpt4}.
Many previous studies exploit quantization methods to reduce LLM inference cost by reducing storage and accelerating computation~\cite{gptq,awq,spqr,squeezellm,smoothquant}.
Quantization with the lower bit-width can further reduce the storage and memory access overheads, leading to a more economic inference. 
However, for 2-bit weight quantization, these methods including Greenbit~\cite{greenbit} still fail to prevent the accuracy loss ($>$3\%).
Compared with single-precision methods, the state-of-the-art methods use mixed-precision methods~\cite{llm-mq,aptq} for LLMs to achieve better accuracy exemplified by the Llama-2 family~\cite{llama2}.



 \begin{figure*}[!t]
  \centering
  \vspace{-10pt}
  \includegraphics[width=0.98\textwidth]{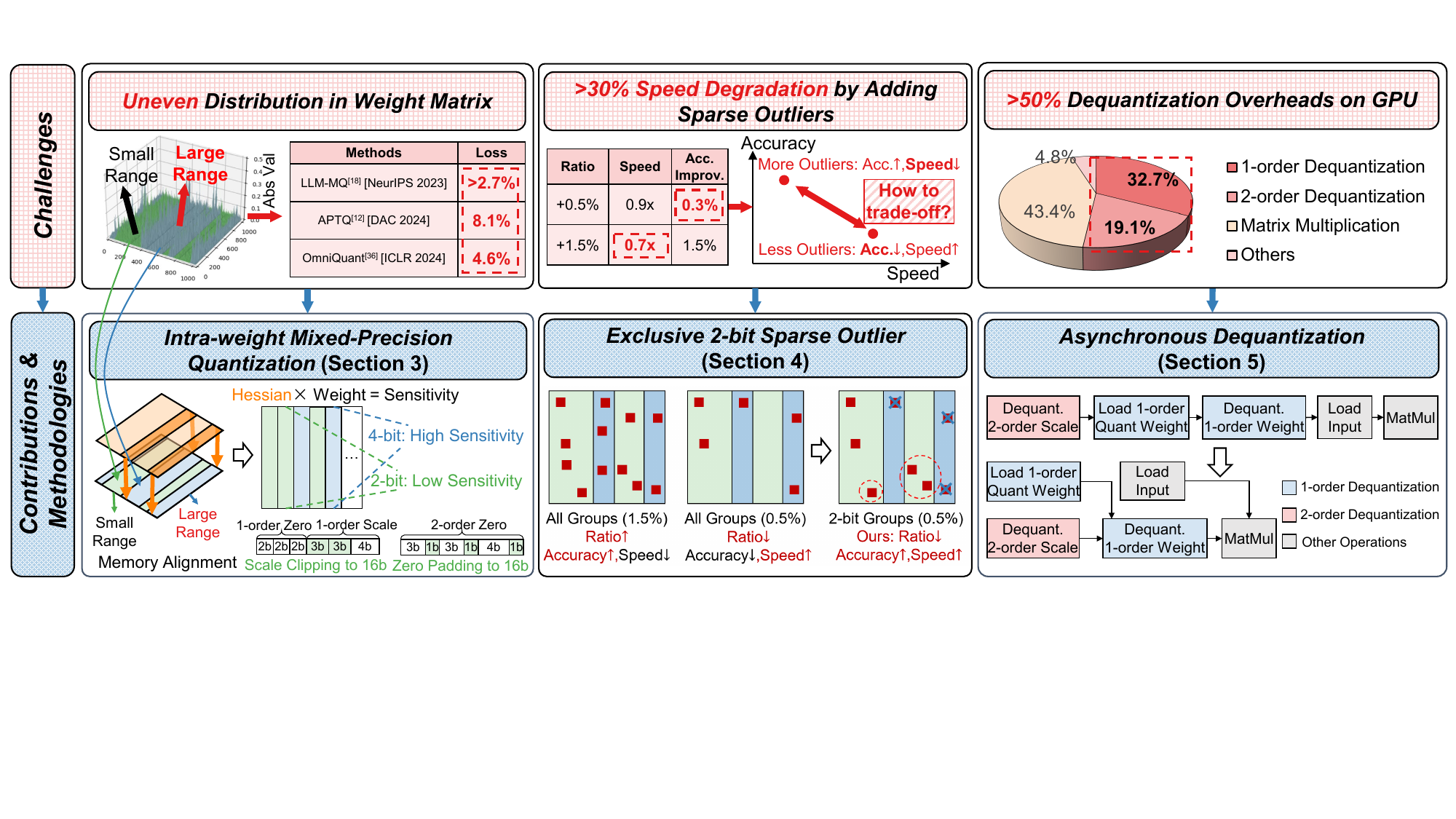}
  \vspace{-13pt}
  \caption{Challenges in fast and efficient quantized LLM inference on GPU: (1) uneven distribution in weight matrix, (2) large speed degradation by adding sparse outliers, and (3) time-consuming dequantization overheads. We propose three novel contributions: (1) intra-matrix mixed-precision quantization, (2) exclusive 2-bit sparse outlier with minimum speed degradation, and (3) asynchronous dequantization, to solve these challenges.}
  \vspace{-15pt}
  \label{fig:overview}
\end{figure*}

However, there still exist challenges of reducing LLM inference cost with the mixed-precision weight quantization method.
\textbf{(1) Uneven distribution in weight matrix.}
Weights are quantized by groups, while some groups contain weights with large range.
Previous works like GPTQ~\cite{gptq}, AWQ~\cite{awq}, LLM-QAT~\cite{llmqat} and Greenbit~\cite{greenbit} quantize weight matrix with single-precision and incur the 3.2\% to 5.6\% accuracy loss for Llama2-7b with 2-bit quantization.
And the recent works apply mixed-precision methods to improve accuracy such as APTQ~\cite{aptq} and LLM-MQ~\cite{llm-mq}. 
They apply inter-matrix mixed-precision quantization and neglect the difference among the groups in each weight, still resulting in $>$2.7\% accuracy loss, as shown in Figure~\ref{fig:overview} left top.
\textbf{(2) Large speed degradation by adding sparse outliers.}
Reserving sparse outliers improves accuracy but introduces speed degradation affected by the outlier ratio.
Figure~\ref{fig:overview} middle top illustrates that the relationship between the speed and accuracy with different sparse outlier ratios (0.5\% and 1.5\%).
SpQR reserves 1.5\% sparse outliers to improve the accuracy, resulting in $>$30\% speed degradation and hindering the inference speedup~\cite{spqr} while reserving less brings limited accuracy improvement.
Consequently, it is necessary to improve accuracy and speed simultaneously by adding a small fraction of sparse outliers in specific regions.
\textbf{(3) Time-consuming dequantization operations on GPUs.}
Mainstream methods require a dequantization operation to perform computation on 2-bit weights.
Furthermore, a 2-order dequantization operation is applied because the scales of groups are also quantized.
Figure~\ref{fig:overview} right top shows that, the dequantization operation takes over 50\% of the total execution time.
These challenges become the crucial factors impeding fast and efficient LLM inference.

In response to these challenges, we enable fast and efficient LLM inference on GPUs with the following contributions in this paper:


\textbf{(1) Intra-matrix mixed-precision quantization.}
We point out that the range of weights by groups varies and these groups always exhibit high sensitivity (large Hessian value and range variation).
Because 4-bit quantization can prevent accuracy loss~\cite{awq,gptq}, we only quantize 25\% of sensitive groups with large Hessian value and range variation using 4-bit, and we also apply scale clipping and zero padding techniques to achieve the memory alignment.
Such a method reduces the accuracy loss for 2-bit quantization from 4.6\% to 2.2\% for Llama2-7b and from 8.6\% to 2.5\% for Llama-7b.

\textbf{(2) Exclusive 2-bit sparse outlier with minimum speed degradation.}
We point out that the speed degradation by reserving sparse outliers consists of launching GPU kernel and calculation. The time of launching kernel is fixed while the calculation time is proportional with sparse outlier ratio.
Thus, we only reserve a small fraction ($<$0.5\% with only launching kernel time) of large weights in 2-bit groups as sparse outliers, which leads to a lower average bit increment and higher speed. Such design improves the accuracy by $>$0.5\% with $<$0.15 increased average weight bit and only 20\% speed degradation.

\textbf{(3) Asynchronous dequantization.}
We point out that calculating the scales of each group in 2-order dequantization is independent of the loading group weights in 1-order dequantization.
Thus, we design the asynchronous dequantization and accelerate the GPU kernel by up to 3.92$\times$.

We conduct extensive experiments on different model families (\textit{e.g.,} Llama1~\cite{llama}, Llama2~\cite{llama2}, Llama3~\cite{llama3}, ChatGLM3~\cite{chatglm3} and BERT~\cite{bert}) and model sizes (\textit{i.e.,} Llama2-7b to Llama2-70b).
We achieve 2.91-bit for each weight considering all scales/zeros with 25\% 4-bit for different models with negligible accuracy loss (\textit{e.g.} 2.2\%/1.1\% for Llama2-7b/13b, respectively).
The end-to-end speedup for Llama2-7b is 1.74$\times$, and we reduce both runtime cost and total cost by up to 2.53$\times$ and 2.29$\times$ with less GPU requirements.

\section{Background}

\begin{figure}[!t]
  \centering
  \includegraphics[width=0.48\textwidth]{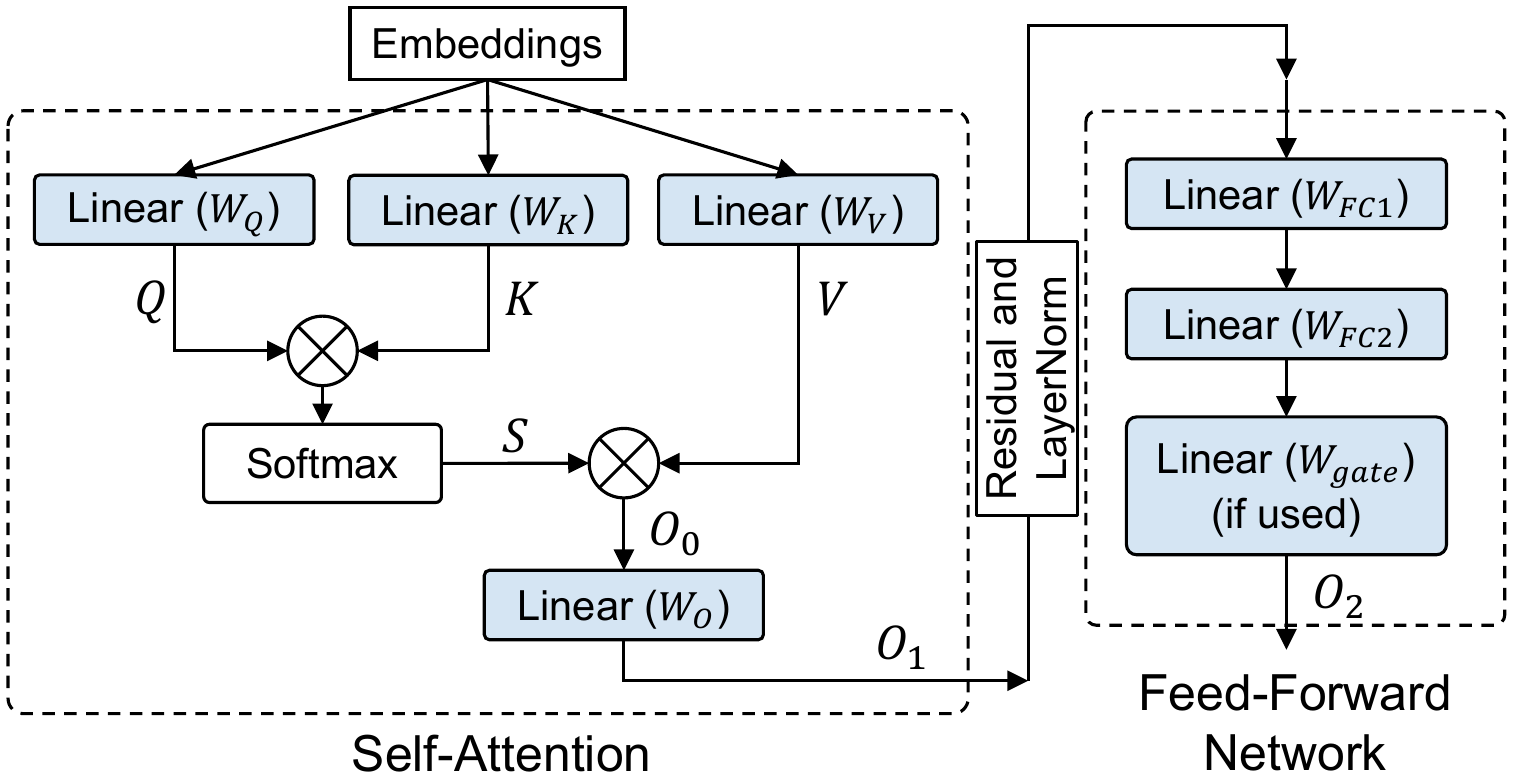}
  \vspace{-20pt}
  \caption{Overview of each transformer block. The self-attention contains four linear projections and feed-forward network contains at least two linear projections.}
  \label{fig:transformer}
  \vspace{-15pt}
\end{figure}

\subsection{Transformer}
The transformer model~\cite{transformer} is a typical backbone architecture primarily used for LLM tasks such as language translation, text summarization, and question answering~\cite{language_translation,text_summarization,question_answer}.
The transformer model consists of several transformer blocks, and each block contains two components, self-attention and feed-forward network.

\textbf{Self-attention:}
The self-attention mechanism in the Transformer model involves four computations.
First, the input embedding $X$ is multiplied by four weights matrix $W_Q$, $W_K$, $W_V$ to get Query, Key, and Value vectors ($Q$, $K$, and $V$) separately. 
Second, the attention scores are computed by taking the dot product of $Q$ with $K$ and scaled down by $\sqrt{d_k}$, the square root of the dimension of $K$.
Then, the scaled attention scores are normalized by a softmax function to obtain a relative attention scores, which allows the model to focus on relevant parts of the input sequence.
And the softmax-normalized attention scores are multiplied by $V$ to obtain the weighted output $O_0$. In addition, $O_0$ is multiplied by $W_O$ to obtain the final output $O_1$ of self-attention.
\begin{equation}
  O_0=Softmax(\frac{QK^T}{\sqrt{d_k}})V
\end{equation}

\textbf{Feed-forward Network:}
After the self-attention, the output $O_1$ is passed through a feed-forward network within each layer. This network applies two or three linear transformations to $O_1$ and obtain the final output $O_2$.

\subsection{Quantization}
A group of floating weights with range of $(w_{min},w_{max})$ are quantized to integers ($w_{int}$) with the range of $(0,2^N-1)$, a zero-point ($z$) and a scaling factor ($s$)~\cite{quantwhite}. 
Scaling factor is half and zero-point is an $N$ bits integer. 
The quantization and dequantization operations are:
\begin{equation}
  s=\frac{w_{max}-w_{min}}{2^N-1}
\end{equation}
\begin{equation}
  z=round(\frac{-w_{min}}{s})
\end{equation}
\begin{equation}
  w_{int}=clamp(0,2^N-1,round(\frac{w}{s})+z)
\end{equation}
\begin{equation}
  w_{float}=(w_{int}-z)\times s
\end{equation}

Weights along the direction of input dimensions are quantized by group, while the scaling factors of groups along the direction of the output dimensions are also quantized to further reduce the average bit.
Here, the average bit, denoted as $\overline{bit}$ with 1-order and 2-order quantization is as follows:
\begin{numcases}{\overline{bit}=}
N+\frac{N+16}{g_1}, & 1-order \\
N+\frac{N+N_s}{g_1}+\frac{N_s+16}{g_1 \times g_2}, & 2-order
\end{numcases}
where $g_i$ represents the size of group and $N_s$ represents the bit width of the quantized scaling factors with 2-order.


\subsection{Sparse Outlier}

In LLMs, weights exhibit large magnitudes compared to the others. 
By applying the normal uniform quantization, the quantization range is expanded by these outliers and thus, the majority of weights are quantized with only a few representations, leading to large quantization error. 
Mixed-precision quantization methods like SpQR, SqueezeLLM and LLM-MQ filter and retain outliers that negatively impact quantization error and store outliers with 16-bit sparse matrix~\cite{spqr,squeezellm,llm-mq}. 
These methods quantize the weights matrix into a dense matrix and a 16-bit sparse matrix. 
And the Compressed Sparse Row (CSR) is used to represent sparse matrices, optimizing memory usage. 

In CSR, three arrays encapsulate the matrix data. 
The values array (values) efficiently arranges non-zero elements in row-major order, while the column indices (col\_ind) array meticulously records the corresponding column index for each non-zero element. 
The row pointer (row\_ptr) array serves as a guide, storing the starting index for each row in the values array. 
Outlier retention of weights is common for low-bit quantization in LLMs. Each outlier is stored using one 32-bit combination: a 16-bit weight value and a corresponding 16-bit column index. 
Additionally, for every row, a 32-bit number is allocated to store the total count of outliers up to that row. 
This results in an overall 32-bit storage overheads for each weight outlier.

\begin{figure}[!b]
  \centering
  \vspace{-20pt}
  \includegraphics[width=0.47\textwidth]{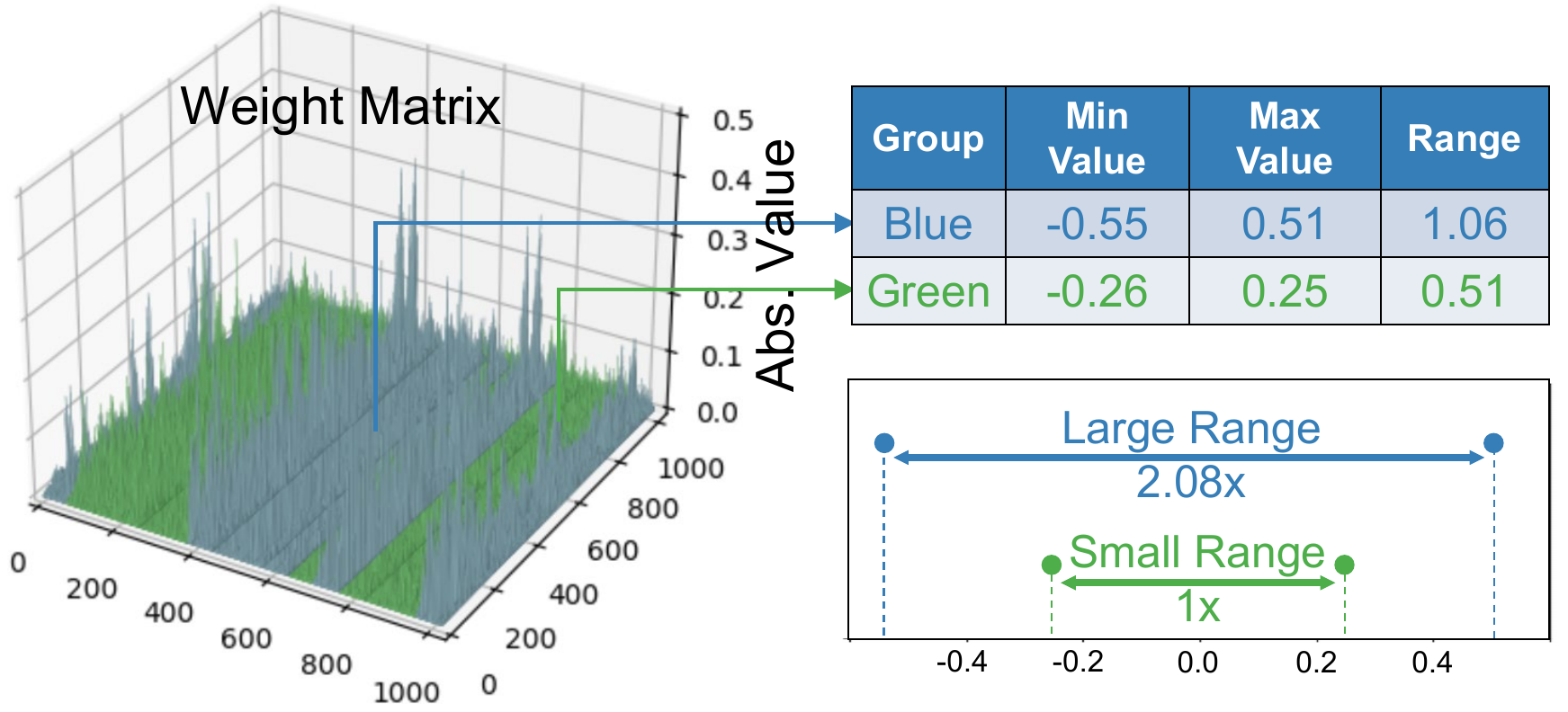}
  \vspace{-16pt}
  \caption{The range of weight matrix by groups: the blue groups have larger range while the green groups are smaller.}
  \label{fig:m1_challenge}
\end{figure}

\begin{figure*}[!t]
  \centering
  \includegraphics[width=0.98\textwidth]{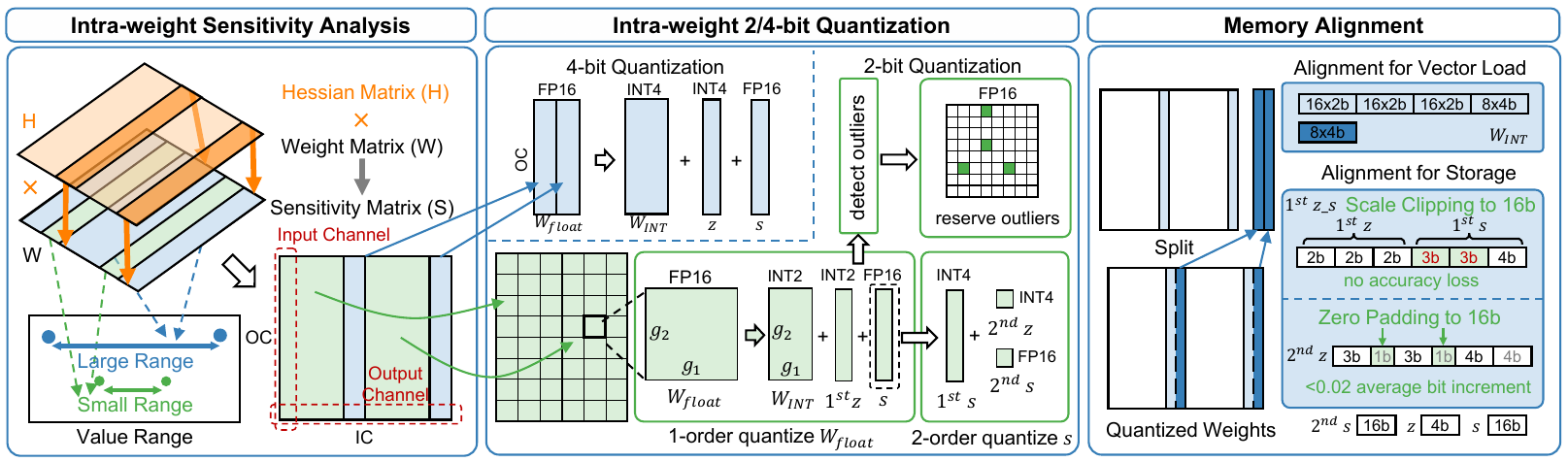}
  \vspace{-12pt}
  \caption{Overview of range-aware quantization method. We first analyze the sensitivity $S$ of different groups by the multiplication of the original weight matrix $W$ with different range distribution and Hessian matrix $H$, and then apply intra-weight 2/4-bit mixed-precision quantization. Last, we propose three techniques for memory alignment.}
  \vspace{-15pt}
  \label{fig:raq}
\end{figure*}

\section{Intra-Weight Mixed-precision Quantization}\label{sec:m1}

The quantization method is widely used to reduce LLM inference cost~\cite{gptq,awq,spqr} by reducing storage and accelerating computation. 
It uses discrete values to represent continuous weights, leading to quantization errors. 
For 2-bit quantization, weights are quantized by groups, while some groups contain weights with large range.

\subsection{Challenge}
Previous works~\cite{gptq,awq,llmqat,greenbit} only quantize weight matrix with 2-bit single-precision and incur the 3.2\% to 5.6\% accuracy loss for Llama2-7b.
The state-of-the-art methods use inter-weight mixed-precision methods~\cite{llm-mq,aptq} for LLMs to improve the accuracy with $<$3 average bit considering all scaling factors and zero-points. 
However, these methods neglect the range variance of quantization groups (uneven distribution) inside each weight matrix, which leads to $>$2.7\% accuracy loss for Llama2-7b.
Thus, the key challenge is that even the state-of-the-art mixed-precision methods suffer from handling the uneven distribution inside weight matrices.

\subsection{Motivation and Insight}
Some previous methods like SpQR~\cite{spqr}, SparseGPT~\cite{sparsegpt} and AWQ~\cite{awq} have analyzed that the quantization errors are large in specific groups.
So we depict the range of weights by groups, as shown in Figure~\ref{fig:m1_challenge}.
The range of weights exhibits different distributions for different groups, and only a small fraction of groups show a large range while the others show a small range.
Applying 4-bit quantization to these groups with a large range can reduce the quantization error. 
Thus, our key insight is, \textbf{the range by groups varies inside weight matrix and some groups require 4-bit quantization to reduce large quantization error.}

\subsection{Approach}
\textbf{Intra-weight Quantization.} 
In Figure~\ref{fig:raq}, we depict the quantization flow of weights. 
Notably, weights are quantized by group to reduce accuracy loss while the groups with a large range still require larger bit-width to improve accuracy. 
To determine which values are more sensitive to affect the loss caused by quantization, we perform Taylor series~\cite{taylor} expansion to analyze how the model output changes in response to perturbations in the parameters $W$:
\begin{equation}
\begin{split}
    L(W_Q) &\approx L(W)-g^T(W-W_Q)+\frac{1}{2}(W-W_Q)^TH(W-W_Q) \\
    &\approx L(W)+\frac{1}{2}(W-W_Q)^TH(W-W_Q)
\end{split}
\end{equation}
where $g$ is the gradient and $H$ is the second derivative (i.e., Hessian matrix) of the loss at $W$. 
Assuming that the model has converged to a local minimum, the gradient $g$ can be approximated as zero, which is mentioned in previous works~\cite{gptq,squeezellm,billm}. The quantization error $W-W_Q$ is weighted by the second-order derivative $H$. 
This highlights the importance of minimizing perturbations for weights that have large Hessian values, as they have a greater impact on the overall perturbation of the final output. 
In other words, the second-order derivative serves as a measure of importance for each weight value.
\textbf{To reduce the accuracy loss, we should consider both the range variation and the second derivative variation by groups.}

Therefore, we first analyze the range variation by group.
The range of groups in the same input channels is merged to streamline the complexity of hardware computation. 
Concretely, weights are divided by the Hessian inverse matrix to get the sensitivity of weights, leveraging the Hessian inverse matrix's ability to discern weight importance. 
Subsequently, the calibrated weights are squared to magnify the sensitivity variance. 
The sensitivity of each input channel is computed as follows:
\vspace{-3pt}
\begin{equation}
    S_i=\sum_{m=1}^{M} \sum_{j=1}^{OC} \frac{w_{Mi+m,j}^2}{[H^{-1}]_{Mi+m}^2}
\end{equation}
where $H^{-1}$ represents the Hessian inverse matrix of each weight matrix, $M$ represents the groupsize, $OC$ represents the dimension of output channels, and $w_{Mi+m,j}$ represents the original value of each element in the group $m$.
The Hessian inverse matrix $H^{-1}$ can be calculated offline~\cite{gptq} to reduce the quantization overheads.
And $S_i$ represents the sensitivity of group $i$, which the serves as a criterion and a indicator to determine which group should be quantized with higher bitwidth.

Thus, the variance of weights by groups is equivalent to the sensitivity variance. 
Then, we analyze the variance and obtain the information for $S_i$ that the groups with large sensitivity require 4-bit quantization and others require 2-bit quantization. 
According to the analysis results, we apply the finetuning approach~\cite{llmqat,omniquant} like quantization aware training (QAT) to further reduce the quantization error. 
In the forward process of QAT, the weights with a large range are quantized and then dequantized with 4-bit by group while others are with 2-bit. 
After finetuning, we use the same analysis results to quantize different grouped input channels with different bit-width. 
For small range input channels, we first do 2-bit 1-order quantization for the weights $W_{float}$ by group $g_1$ in the direction of input channels to get the quantized weight $W_{INT}$, 1-order zeros $1^{st} z$ and scales $s$. 
To further reduce the average bit, the scales $s$ are also quantized with 4-bit by group $g_2$ in the direction of output channels (2-order quantization) to get 1-order quantized scales $1^{st}s$, 2-order zeros $2^{nd}z$ and scales $2^{nd}s$. 
During 2-bit quantization, outliers are also detected and reserved to reduce accuracy loss. 
Then, the weights in large range input channels are quantized with 4-bit.
Therefore, the average bit-width of each weight element considering all scaling factors and zero-points is:
\vspace{-3pt}
\begin{equation}
    \overline{bit}=\alpha\times(2+\frac{2+N_s}{g_1}+\frac{N_s+16}{g_1g_2})+(1-\alpha)\times4
\end{equation}
where $\alpha$ is the ratio of 2-bit and other parameters are as the same definition in equation (6) and (7).

\textbf{Memory Alignment.} 
As illustrated in Figure~\ref{fig:raq} right, the quantized weights are split into two quantized weight matrixes. 
The first matrix stores 3 groups of 2-bit quantized weights (the size of group is 16) and 8 4-bit quantized weights into continuous memory. 
The other matrix stores 8 remaining 4-bit quantized weights into independent memory. 
Thus, the memory access for each thread in GPU is memory-aligned.
Then, to maximize memory utilization without compromising accuracy, a nuanced approach is taken with scale factors and zero values. 
1-order zeros $1^{st} z$ and scales $1^{st} s$ are coherently packed together and two of the three scales are compressed into 3-bit representations to occupy a 16-bit memory space efficiently (Scale Clipping). 
In contrast, 2-order zeros are stored with paddings (Zero Padding) to maintain alignment with memory boundaries. 
These meticulous memory alignment techniques effectively preserve accuracy loss while incurring a negligible <0.02 average bit overhead.

\section{Exclusive 2-bit Sparse Outlier}
The intra-weight mixed-precision quantization method can efficiently reduce quantization errors of groups with large variations. However, there are still some outliers sparsely distributed in the weight matrices. 
Reserving these sparse outliers can further improve accuracy but it also introduces speed degradation affected by the sparse outlier ratio.
\begin{figure}[!t]
  \centering
  \includegraphics[width=0.48\textwidth]{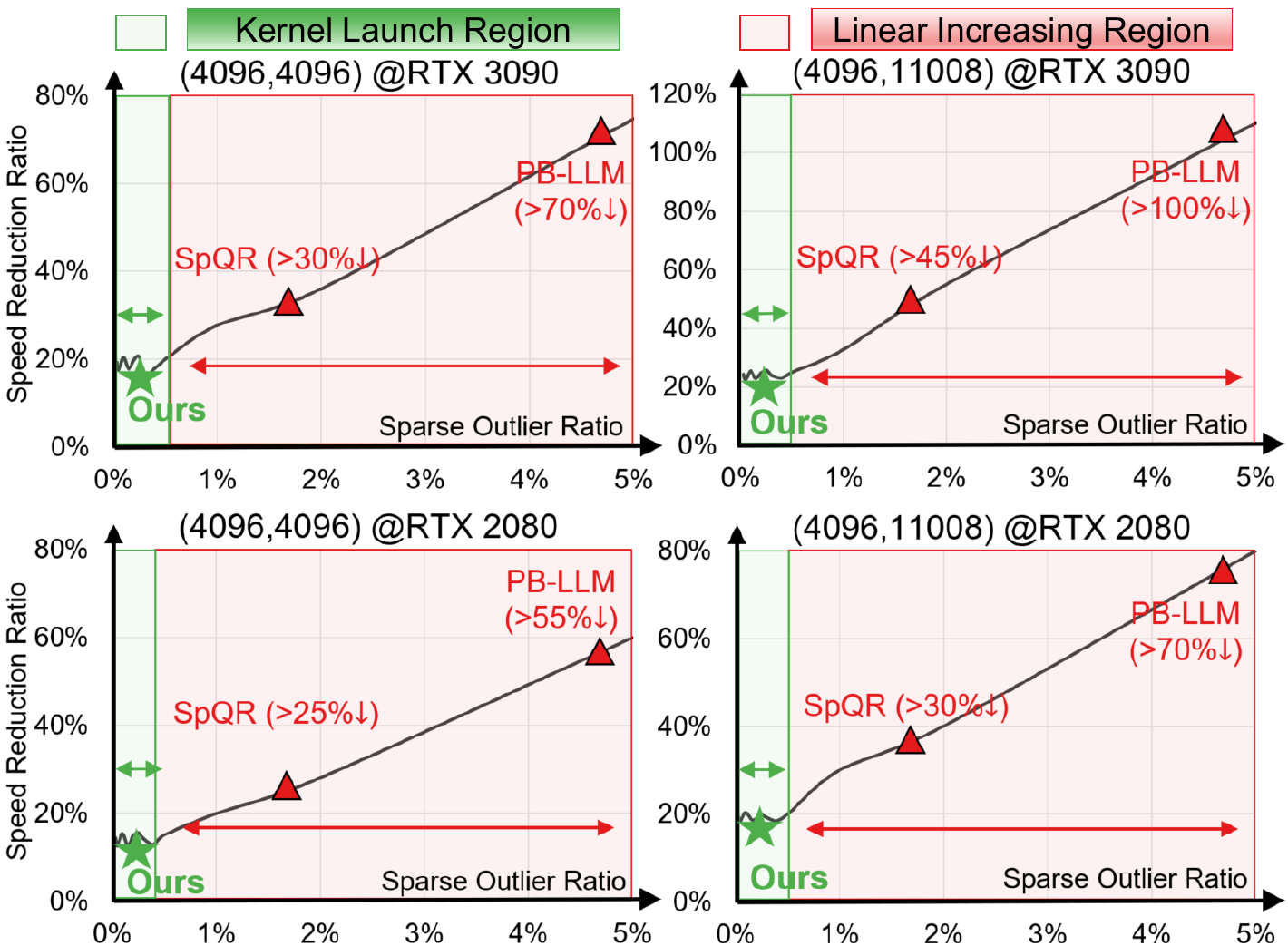}
  \vspace{-20pt}
  \caption{The GPU kernel speed degradation ratio of SpMV+GEMV with dequantization compared to GEMV with dequantization with different shape on different GPUs. Kernel launch region is irrelevant with sparse outlier ratio. Linear increasing region is increased linearly with the ratio.}
  \vspace{-10pt}
  \label{fig:m2_challenge}
\end{figure}

\subsection{Challenge}
Reserving these sparse outliers can further improve accuracy but it also introduces speed degradation affected by the outlier ratio.
In order to reduce the quantization error caused by these random outliers, the straightforward approach is to reserve a large amount of outliers from all groups in weight matrix, which introduces speed degradation and large average bit increment. 
Previous works reserve 1.5\% sparse outliers to improve the accuracy, and apply a sparse matrix-vector multiplication (SpMV) to compute these outliers.
Due to the differences in computational processes between the SpMV and the general matrix-vector multiplication (GEMV) with dequantization, these two parts need to be calculated separately and then combined at then end, resulting in $>$30\% speed degradation and hindering the inference speedup~\cite{spqr}. 
And reserving less outliers maintains the speed but brings limited accuracy improvement.
Thus, the key challenge is to maximize the accuracy improvement and maintain speed simultaneously by reserving sparse outliers.



\subsection{Motivation and Insight} 
We depict the GPU kernel speed degradation ratio caused by sparse outliers in Figure~\ref{fig:m2_challenge}.
In kernel launch region, the speed degradation ratio is fixed while in linear increasing region it is increased linearly with sparse outlier ratio.
Besides straightforwardly reserving a large ratio of outliers from all groups in the weight matrix, another way is to reserve a small ratio of outliers from some specific groups.
Previous designs like SpQR~\cite{spqr} and SqueezeLLM~\cite{squeezellm} apply the corresponding sparse format (\textit{e.g.,} CSR) to store and compute the sparse outliers. 
Each sparse outlier requires at least one 16-bit for the weight representation and one 16-bit for the position.
Therefore, besides the speed degradation, introducing 1.5\% 16-bit sparse outliers leads to $(16+16-2)\times1.5\%=0.45$ extra average bit.
Thus, our key insight is, \textbf{reserving a small fraction (\textit{e.g.,} $\leq$0.5\%) of sparse outliers from 2-bit groups can maintain inference speed and the algorithm accuracy simultaneously}. 

\begin{figure}[!b]
  \centering
  \vspace{-15pt}
  \includegraphics[width=0.48\textwidth]{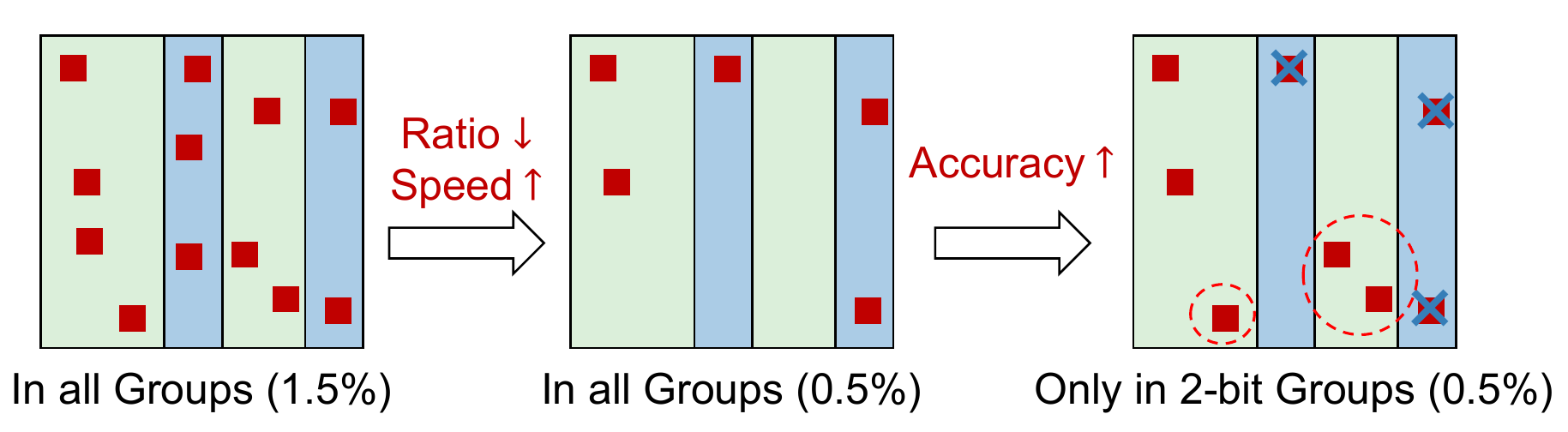}
  \vspace{-20pt}
  \caption{Exclusive 2-bit sparse outlier reservation method. The green regions represent 2-bit groups while the blue regions represent 4-bit groups. The red squares represent sparse outliers. We constrain the ratio of sparse outliers from 1.5\% to 0.5\% and from all groups to only 2-bit groups.}
  \label{fig:m2_overview}
\end{figure}

\subsection{Approach}
The detailed method of the exclusive 2-bit sparse outlier is illustrated in Figure~\ref{fig:m2_overview}.
The green regions represent weights quantized to 2-bit precision, while the blue regions represent weights quantized to 4-bit precision.
The red squares represent sparse outliers.
The original ratio of outliers is 1.5\% and they are reserved from all groups including 2-bit groups and 4-bit groups.
First, we constrain the ratio of sparse outliers to 0.5\%, which decreases the speed degradation ratio from $>$30\% to the minimum (about 20\%, for GEMV with the size of (4096,4096) on NVIDIA RTX 3090).
Then, we constrain the reservation regions from all groups to only 2-bit groups. 
Thus, with the minimum inference overheads, our method reserves more outliers in 2-bit groups.
In practice, we only reserve 0.2-0.5\% outliers and increase 0.06-0.15 average bit. 
We utilize the CSR format used in previous designs~\cite{spqr,squeezellm} for sparse outliers representation, and apply GPU kernels in the NVIDIA cuSPARSE library~\cite{cuSPARSE} to perform matrix-vector multiplication on these outliers.

\begin{algorithm}[!t]
\small
    \renewcommand{\algorithmicrequire}{\textbf{Input:}}
    \renewcommand{\algorithmicensure}{\textbf{Output:}}
    \caption{GEMV with Asynchronous Dequantization GPU Kernel Algorithm}\label{gemv kernel}
    \begin{algorithmic}[1]
        \REQUIRE Quantized weights $W_{int}$, 1-order zero\_points $1^{st}Z$, 1-order scaling\_factors $1^{st}S_{int}$, 2-order zero\_points $2^{nd}Z$, 2-order scaling\_factors $2^{nd}S$, input $X$, input channel $IC$.
        \STATE /*Hyperparameters for 2-bit dequantization*/
        \STATE $N_{worker}=2048$, $N_{pack}=16$ 
        \STATE $L_o=Ceil(IC/N_{worker})$
        \STATE $L_i=4$ /*Equal with vector load \texttt{double4}*/
        \FOR{\_ from $1$ to $L_o$}
            \STATE /*load $1^{st}S_{int}$, $2^{nd}Z$, $2^{nd}S$, $X$ to Shared Memory*/
            \STATE $sm\_s\_1st \gets 1^{st}S_{int}$, $sm\_x \gets X$
            \STATE $sm\_z\_2nd \gets 2^{nd}Z$, $sm\_s\_2nd \gets 2^{nd}S$
            \STATE \_\_$syncthreads()$
            \FOR{\_ from $1$ to $L_i$}
                \STATE /*load $W_{int}$, $1^{st}Z$ to Shared Memory*/
                \STATE $sm\_w\_int \gets W_{int}$, $sm\_z\_1st \gets 1^{st}Z$ 
                \STATE /*load to registers*/
                \STATE $1^{st}s_{int} \gets sm\_s\_1st$, $2^{nd}z\gets sm\_z\_2nd$, $2^{nd}s\gets sm\_s\_2nd$
                \STATE $1^{st}s \gets dequantize(1^{st}s_{int}, 2^{nd}z,2^{nd}s)$
                \STATE \_\_$syncthreads()$
                \FOR{\_ from $1$ to $N_{pack}$}
                    \STATE /*load input to register*/
                    \STATE $w_{int}\gets sm\_w\_int$, $1^{st}z\gets sm\_z\_1st$, $x\gets sm\_x$
                    \STATE $w_{float} \gets dequantize(w_{int}, 1^{st}z, 1^{st}s)$
                    \STATE $psum \gets psum + w_{float} \times x$
                \ENDFOR
            \ENDFOR
        \ENDFOR
        \STATE $parallel\_reduce(O, psum)$
        \ENSURE Output vector $O$.
    \end{algorithmic}
\end{algorithm}

\section{Asynchronous Dequantization}

\subsection{Challenge}
Because the weights are quantized by 2/4-bit intra-weight mixed-precision, it requires the dequantization operation to restore the weights to half data type (16-bit) before performing the multiplication between the input and weights. 
We apply 1-order and 2-order quantization to quantize the weights of each group and the scales of adjacent groups, respectively. Previous designs (\textit{e.g.}, SpQR\cite{spqr}, Greenbit\cite{greenbit}) use the synchronous dataflow (\textit{i.e.}, performing dequantization after loading all weights), resulting in $>$50\% overheads of end-to-end execution time. 
Thus, the key challenge is the time-consuming synchronous dequantization operation becomes the bottleneck in accelerating LLM inference after quantization. 

\subsection{Motivation and Insight}
The 1-order and 2-order quantization requires two synchronous dequantizations in dataflow. 
However, the 2-order dequantization for calculating the scales of each group is independent of the weights of each group for the 1-order dequantization. Moreover, the 1-order dequantization for restoring the weights to 16-bit is independent of the input data of multiplication. 
Thus, our key insight is \textbf{calculating the scales of each group in 2-order dequantization can be overlapped with loading weights of each group in 1-order dequantization in GPU kernel.}

\begin{figure}[!t]
    \centering
    \includegraphics[width=0.49\textwidth]{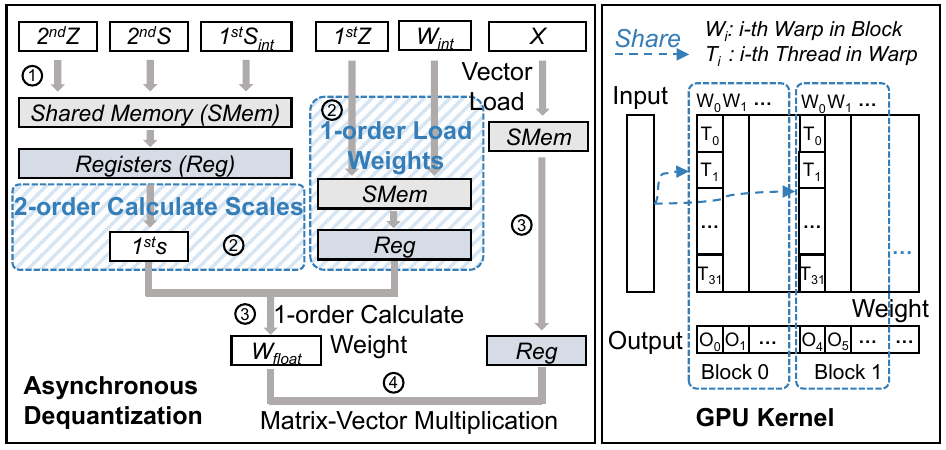}
    \vspace{-20pt}
    \caption{Asynchronous dequantization on GPUs.}
    \vspace{-15pt}
    \label{fig:m3_gemv}
\end{figure}

\subsection{Approach} 
Based on the insights above, we design the asynchronous dequantization dataflow as illustrated in Figure~\ref{fig:m3_gemv} on GPUs. 
With the help of the shared memory, we can overlap the calculating scales of each group and the loading weights of each group. 
Further, we use CUDA primitive \texttt{\_\_shfl\_down\_sync()} to reduce the partial result inside a warp in parallel. 
From the perspective of memory access, we apply vector load (\textit{e.g.} \texttt{double4}) technique to load quantized weights, input, scales and zeros to minimize the numbers of memory access.
Algorithm~\ref{gemv kernel} shows the four main parts including outer loop, inner loop, packed loop and parallel reduction. 
The outer loop (Line 5-24) loads the 2-order zero-points $2^{nd}Z$, quantized 1-order scaling factors $1^{st}S_{int}$ and 2-order scaling factors $2^{nd}S$ from global memory to shared memory. 
Then, it prepares for the 2-order dequantization, corresponding to \ding{192} in Figure~\ref{fig:m3_gemv}.
The inner loop (Line 10-23) loads the quantized weight $W_{int}$ and the 1-order zero-points $1^{st}Z$ from global memory to shared memory.
Simultaneously, it also performs 2-order dequantization for scaling factors $1^{st}s$, corresponding to \ding{193}.
The packed loop (Line 17-22) uses the dequantized scaling factors $1^{st}s$, 1-order zeros-point $2^{nd}Z$ and quantized weights $W_{int}$ to perform 1-order dequantization, and then loads input vector $X$ for multiplication, corresponding to operation \ding{194} and \ding{195}.
Last, parallel reduction is performed to add partial results from all threads.

\section{Experimental Result}
\subsection{Experimental Setup}\label{sec:setup}
\textbf{Benchmarks.} 
We conduct comprehensive experiments on the Llama1~\cite{llama}, Llama2~\cite{llama2}, Llama3~\cite{llama3} model families and the ChatGLM3-6b model~\cite{chatglm3}, which are owing to critical and efficient influence in recent model advancements. 
We focus on two primary metrics: \textit{perplexity} and \textit{zero-shot performance}.
The perplexity (PPL) is evaluated by the WikiText-2~\cite{wikitext} benchmark.
The zero-shot performance is assessed across four zero-shot benchmarks, namely Piqa~\cite{piqa}, HellaSwag~\cite{hellaswag}, WinoGrande~\cite{winogrande}, and Arc-e~\cite{arc-ec}. 
We also conduct experiments on the BERT-base~\cite{bert} model on two datasets (MNLI and STS-B) of the GLUE benchmark~\cite{glue}.

\textbf{Baselines.} 
We compare our method with state-of-the-art single-precision and mixed-precision quantization methods, including AWQ~\cite{awq}, GPTQ~\cite{gptq}, OmniQuant~\cite{omniquant}, APTQ~\cite{aptq}, LLM-MQ~\cite{llm-mq} and Greenbit~\cite{greenbit}. 
For kernel and end-to-end performance, the original PyTorch implementation on HuggingFace~\cite{HuggingFace} is also used as the baseline.

\textbf{Hardware Platforms.} 
We implement our design and compare with other baselines on NVIDIA RTX 2080, NVIDIA RTX 3090, and NVIDIA A100 (80G) GPUs with CUDA version 12.2. 
As mentioned in Figure~\ref{fig:cost_comparison} and Table~\ref{tab:llm cost}, the runtime cost is evaluated considering the power consumption using the NVML library~\cite{nvml}, and the hardware buying cost is evaluated considering the price (\$12,500 for A100, \$1499 for RTX 3090, and \$699 for RTX 2080). With the 5-year service life period, the buying cost for each GPU is: \$0.285/h for A100, \$0.034/h for RTX 3090, and \$0.016/h for RTX 2080. We assume the electricity cost is 0.09\$/kWh~\cite{electricity_cost}, and the total cost for each GPU can be calculated by:

\vspace{-12pt}
\begin{equation}
\begin{aligned}
        buying\ cost &= price \div 5\ years \\
        runtime\ cost &= electricity\ cost \times measured\ runtime\ power \\
        total\ cost &= buying\ cost + runtime\ cost
\end{aligned}
\end{equation}
\vspace{-12pt}


\begin{table}[!t]
\small
    \caption{LLM Algorithm Perplexity and Zero-shot Accuracy}
    \centering
    \vspace{-12pt}
    \label{tab:accuracy}
    \begin{tabular}{lcccc}
    \toprule
        Method & $\overline{bit}$ & PPL($\downarrow$) & Piqa/Hella./Wino./Arc-e & Avg.($\uparrow$) \\ 
        \midrule
        Llama1-7b & 16 & 5.68 & 79.2/76.2/69.9/72.8 & 75.1 \\
        APTQ\cite{aptq} & 3.13 & 6.76 & 74.5/68.3/65.3/57.9 & 66.5 \\
        \textbf{Our} & \textbf{2.84} & \textbf{6.61} & 76.6/72.8/68.5/70.4 & \textbf{72.1}\\ 
        \textbf{Our (0.2\%)} & \textbf{2.91} & \textbf{6.56} & 76.8/73.8/68.7/71.0 & \textbf{72.6} \\ 
        \midrule
        Llama1-13b & 16 & 5.09 & 80.3/79.0/72.7/74.8 & 76.7 \\
        APTQ\cite{aptq} & 3.13 & / & 74.4/71.2/68.0/64.1 & 69.4 \\ 
        \textbf{Our} & \textbf{2.84} & \textbf{5.92} & 79.2/76.9/71.6/72.7 & \textbf{75.1} \\ 
        \textbf{Our (0.2\%)} & \textbf{2.91} & \textbf{5.89} & 79.3/77.8/71.9/73.0 & \textbf{75.5}  \\ 
        \midrule
        Llama2-7b & 16 & 5.47 & 78.5/56.7/67.2/69.3 & 67.9 \\ 
        GPTQ\cite{gptq} & 3 & 8.37 & 71.7/48.2/61.2/56.1 & 59.3 \\ 
        OmniQuant\cite{omniquant} & 3 & 6.65 & 74.1/51.9/63.5/63.8 & 63.3\\ 
        LLM-MQ\cite{llm-mq} & 2.91 & / & 76.5/53.2/65.0/65.8 & 65.1 \\ 
        Greenbit\cite{greenbit} & 2.91 & \textbf{6.09} & 77.2/53.8/65.8/62.5 & 64.8\\ 
        \textbf{Our} & \textbf{2.84} & 6.62 & 75.9/51.3/66.4/66.9 & \textbf{65.1} \\ 
        \textbf{Our (0.2\%)} & \textbf{2.91} & 6.59 & 76.1/52.0/66.8/67.2 & \textbf{65.7} \\ 
        \midrule
        Llama2-13b & 16 & 4.88 & 78.3/59.7/69.6/73.3 & 70.2 \\ 
        GPTQ\cite{gptq} & 3 & 6.44 & 75.2/56.1/64.3/64.2 & 64.0 \\ 
        OmniQuant\cite{omniquant} & 3 & 5.59 & 77.2/56.8/66.5/66.7 & 66.8 \\ 
        LLM-MQ\cite{llm-mq} & 2.91 & / & 77.6/56.9/68.0/72.6 & 68.8\\ 
        \textbf{Our} & \textbf{2.84} & \textbf{5.48} & 77.2/57.6/68.5/71.2 & 68.7 \\ 
        \textbf{Our (0.2\%)} & \textbf{2.91} & \textbf{5.37} & 77.8/58.8/68.6/71.3 & \textbf{69.1}\\ 
        \midrule
        Llama3-8b & 16 & 6.10 & 79.9/60.2/72.8/80.1 & 73.3 \\
        GPTQ\cite{gptq} & 3 & 13.03 & 60.8/41.8/60.9/38.8 & 50.6\\ 
        \textbf{Our} & \textbf{2.84} & \textbf{8.54} & 75.3/54.9/70.3/69.4 & \textbf{67.5}\\ 
        \textbf{Our (0.2\%)} & \textbf{2.91} & \textbf{8.46} & 76.8/55.4/70.5/70.3 & \textbf{68.3} \\ 
        \midrule
        ChatGLM3-6b & 16 & / & 70.8/49.4/61.3/51.1 & 58.2 \\ 
        \textbf{Our} & \textbf{2.91} & / & 66.7/44.7/59.4/48.5 & \textbf{54.8} \\ 
        \bottomrule
    \end{tabular}
\vspace{-15pt}
\end{table}

\subsection{Perplexity and Accuracy Evaluation}
Table~\ref{tab:accuracy} shows algorithm perplexity and accuracy with zero-shot performance on Llama1, Llama2, Llama3 families and ChatGLM3-6b. 
We set the 1-order group size to 16 and 2-order group size to 16 for Llama1-7b/13b/70b, Llama2-7b/13b and Llama3-8b with 0.2\% sparse outliers, and the 1-order group size to 8 and 2-order group size to 16 for and ChatGLM3-6b. 
For Llama1 family, we only compare with APTQ because its code is not open-source and it only provides the results for these LLMs. 
And with the emergent abilities~\cite{emergent} of LLMs, we can achieve lower accuracy loss for Llama2-70b model. Table~\ref{tab:bert_acc} shows the algorithm accuracy on BERT-base.

\textbf{Llama1.} 
Compared with APTQ~\cite{aptq}, we achieve 5.6\% and 5.7\% higher accuracy with lower 2.84 average bit on 7b and 13b models, respectively. 
By reserving 0.2\% sparse outliers, we achieve 5.56\% and 5.7\% higher accuracy with lower 2.91 average bit, respectively.

\textbf{Llama2.} 
Compared with GPTQ~\cite{gptq}, our design significantly reduces the accuracy loss from 8.7\% to 2.2\% for Llama2-7b, and from 6.4\% to 1.1\% Llama2-13b with lower average bit. 
Compared with OmniQuant~\cite{omniquant}, our design significantly reduces the accuracy loss from 4.6\% to 2.2\% for Llama2-7b, and from 3.4\% to 1.1\% Llama2-13b.
Compared with LLM-MQ~\cite{llm-mq}, we achieve 0.6\% and 0.3\% higher accuracy for Llama2-7b and Llama2-13b, respectively. 
Compared with Greenbit~\cite{greenbit} on Llama2-7b, we achieve 0.9\% higher accuracy with the same 2.91 average bit. 
And due to the emergent abilities~\cite{schaeffer2024emergent}, Llama2-70b can prevent more accuracy loss than Llama2-13b.

\textbf{Llama3.} 
On Llama3-8b, compared with GPTQ, our design reduces the accuracy loss from 22.7\% to 5.0\% with lower average bit. 

\textbf{ChatGLM3.}
On ChatGLM3-6b, the accuracy loss 3.4\% is also controllable and it can be lower if the bilingual dataset is used for calibration.

\textbf{BERT.}
On BERT-base model, compared with GOBO~\cite{gobo} and Q-BERT~\cite{qbert} methods, we achieve 2.4\% and 0.4\% higher accuracy with lower 2.85 average bit. 
Compared with the original full precision model, the accuracy loss is only 0.8\%.

\begin{table}[!t]
\small
    \caption{BERT Algorithm Accuracy and Compression Ratio}
    \centering
    \vspace{-12pt}
    \label{tab:bert_acc}
    \begin{tabular}{lccccc}
    \toprule
        Method & $\overline{bit}$ & CR & MNLI & STS-B & Avg.($\uparrow$)\\ 
        \midrule
        BERT$_{base}$ & 32 & 1$\times$ & 84.9 & 89.7 & 87.3 \\
        GOBO\cite{gobo} & 3.26 & 9.8$\times$ & 83.8 & 88.3 & 86.1 \\
        Q-BERT\cite{qbert} & 3.52 & 9.1$\times$ & 81.8 & / & / \\
        \textbf{Our} & \textbf{2.85} & \textbf{11.2$\times$} & \textbf{84.2} & \textbf{88.7} & \textbf{86.5}\\
        \bottomrule
    \end{tabular}
    \vspace{-15pt}
\end{table}

\begin{figure}[!b]
    \centering
    \vspace{-15pt}
    \includegraphics[width=0.47\textwidth]{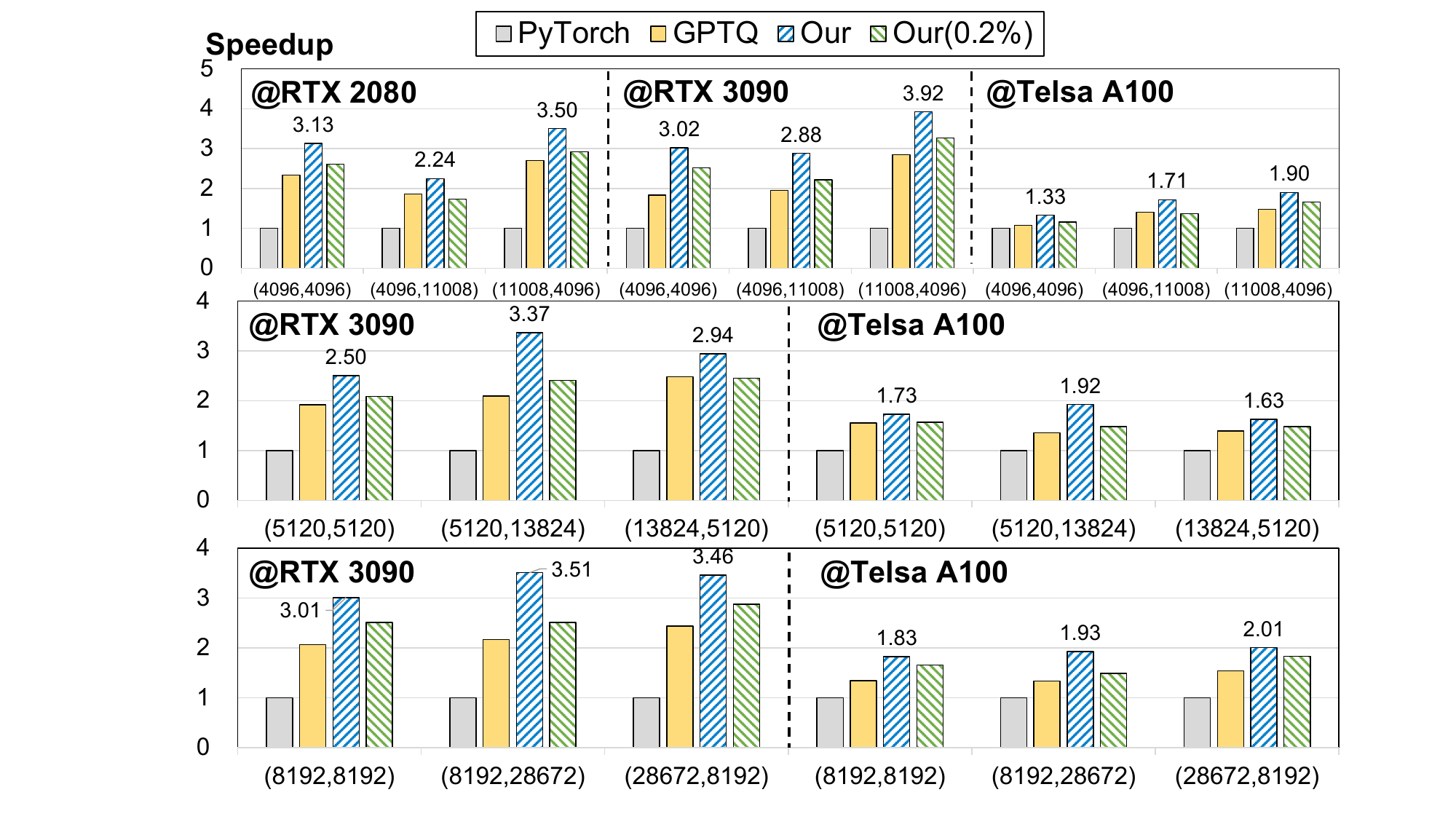}
    \vspace{-12pt}
    \caption{GEMV kernel with dequantization speedup. The tuple in the x-label represents the matrix shape of GEMV kernel.}
    \label{fig:kernel}
\end{figure}

\begin{table*}[!t]
  \caption{Comparison of end-to-end speedup, runtime cost, and total cost}
  \centering
  \vspace{-12pt}
  \label{tab:llm cost}
  \begin{tabular}{cccccccc}
    \toprule
    Model & Method & GPU & Power (W) & Token/s & End-to-end speedup & Runtime cost reduction & Total cost reduction\\
    \midrule
    \multirow{4}{*}{Llama2-7b}
    &PyTorch & 3090$\times$1 & 290/350 & 25.9 & 1$\times$ & 1$\times$ & 1$\times$\\
    \multirow{4}{*}{}&GPTQ & 3090$\times$1 & 320/350 & 37.5 & 1.45$\times$ & 1.31$\times$ & 1.12$\times$\\
    \multirow{4}{*}{}&Our & 3090$\times$1 & 350/350 & 45.2 & \textbf{1.74$\times$} & 1.44$\times$ & 1.16$\times$\\
    \multirow{4}{*}{}&Our & 2080$\times$1 & 150/215 & 34.0 & 1.31x & \textbf{2.53$\times$} & \textbf{2.29$\times$}\\
    \midrule
    \multirow{3}{*}{Llama2-13b}&PyTorch & 3090$\times$2 & 240/350 & 22.0 & 1$\times$ & 1$\times$ & 1$\times$ \\
    \multirow{3}{*}{}&GPTQ & 3090$\times$1 & 270/350 & 23.1 & 1.06$\times$ & 1.89$\times$ & 1.95$\times$\\
    \multirow{3}{*}{}&Our & 3090$\times$1 & 250/350 & 25.0 & \textbf{1.14$\times$} & \textbf{2.19$\times$} & \textbf{2.07$\times$}\\
    \midrule
     &PyTorch & A100$\times$2 & 400/400 & 38.5 & 1$\times$ & 1$\times$ & 1$\times$ \\
    Llama2-70b&GPTQ & A100$\times$1 & 400/400 & 46.5 & 1.21$\times$ & 2.42$\times$ & 2.04$\times$\\
    &Our & A100$\times$1 & 400/400 & 50.5 & \textbf{1.31$\times$} & \textbf{2.62$\times$} & \textbf{2.06$\times$}\\
    \midrule
    \multirow{3}{*}{Llama3-8b}&PyTorch & 3090$\times$1 & 290/350 & 26.5 & 1$\times$ & 1$\times$ & 1$\times$ \\
    \multirow{3}{*}{}&Our & 3090$\times$1 & 350/350 & 48.2 & \textbf{1.82$\times$} & 1.51$\times$ & 1.22$\times$\\
    \multirow{3}{*}{}&Our & 2080$\times$1 & 150/215 & 34.0 & 1.30$\times$ & \textbf{2.61$\times$} & \textbf{2.36$\times$}\\
    \midrule
    &PyTorch & 3090$\times$1 & 350/350 & 32.5 & 1$\times$ & 1$\times$ & 1$\times$ \\
    ChatGLM3-6b&Our & 3090$\times$1 & 350/350 & 40.0 & 1.23$\times$ & 1.23$\times$ & 1.09$\times$ \\
    &Our & 2080$\times$1 & 170/215 & 45.5 & \textbf{1.40$\times$} & \textbf{2.88$\times$} & \textbf{2.44$\times$} \\
  \bottomrule
\end{tabular}
\vspace{-10pt}
\end{table*}

\subsection{Kernel Evaluation}
The detailed speedup of the GEMV kernel with dequantization is shown in Figure~\ref{fig:kernel}. 
We compare our kernel performance with some mainstream quantized kernels (\textit{e.g.}, GPTQ~\cite{gptq}) and the original FP16 models in common LLM GEMV cases.
Because the large accuracy loss of AWQ method with 3-bit quantization, we exclude the comparison with its kernel.
Without computating sparse outliers, our kernel achieves $1.33\times$ $\sim$ $3.92\times$ over \texttt{FP16} on various NVIDIA GPUs.
With 0.2\% sparse outliers, our kernel achieves $1.16\times$ $\sim$ $3.23\times$ over \texttt{FP16} on various NVIDIA GPUs.

\begin{figure}[!b]
    \centering
    \vspace{-15pt}
    \includegraphics[width=0.41\textwidth]{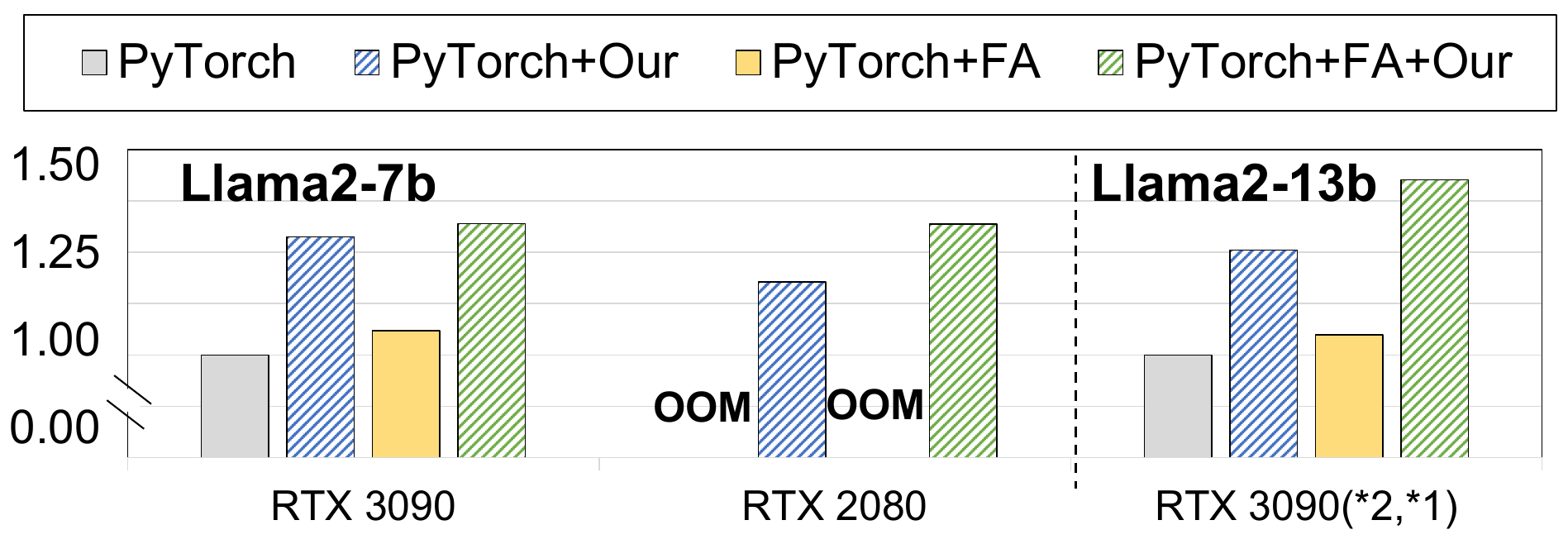}
    \vspace{-10pt}
    \caption{FlashAttention (FA)~\cite{flashattention} integration and comparison.}
    \label{fig:fa}
\end{figure}

\subsection{Performance Evaluation}
We compare the inference cost (end-to-end speedup, runtime cost reduction, and total cost reduction) including 0.2\% sparse outliers among our method, PyTorch, and GPTQ in various model families and model sizes in Table \ref{tab:llm cost}. 
All results are normalized to PyTorch. 
In the Llama2 model family, our design achieves $1.14 \times \sim 1.74 \times$ end-to-end speedup and far outperforms other previous works (\textit{e.g.}, GPTQ~\cite{gptq}) in terms of running cost reduction, reaching $2.19\times \sim 2.53 \times$ over the original model. 
The performance of Llama1 models is the same as Llama2 because they have the same model parameter. 
Moreover, the hardware requirement is dramatically lowered in our design for all models in Table \ref{tab:llm cost}, bringing up to $2.88\times$ runtime cost reduction and $2.44\times$ total cost reduction. 
The performance underscores the effectiveness of our design's optimizations and the practical utility of our approach tailored to optimize the inference and deployment of LLMs.

We also show that our method is also compatible with FlashAttention~\cite{flashattention}, the most widely used LLM inference engine. 
Figure~\ref{fig:fa} shows that our method achieves higher speedup compared with adopting FlashAttention. 
Integrating our method with FlashAttention further accelerates LLM inference by up to $1.31\times$, and even enables LLM inference on GPUs with limited memory (\textit{e.g.,} NVIDIA RTX 2080).

\subsection{Ablation Study}
We also present an ablation study to validate the superiority of our method over manual inter-layer quantization schemes. 
The most intuitive mixed-precision quantization strategy is to uniformly quantize all layers within each block with single-precision. 
The results in Table~\ref{tab:ablation} reveal our method’s efficacy over manual inter-layer quantization for Llama2-7B on WikiText-2, reflected in its consistently lower perplexity across various quantization ratios.

\begin{table}[!h]
\small
    \vspace{-8pt}
    \caption{Ablation study: Perplexity comparison on Llama2-7b}
    \centering
    \vspace{-10pt}
    \label{tab:ablation}
    \begin{tabular}{lccc}
    \toprule
        Method & $\overline{bit}$ & Ratio of 4-bit & Perplexity ($\downarrow$)\\ 
        \midrule
        Manual inter-layer & 2.85 & 25\% & 7.13  \\
        \textbf{Our} & 2.85 & 25\% & 6.62 \\
        Manual inter-layer & 2.60 & 10\% & 9.32 \\
        \textbf{Our} & 2.60 & 10\% & 7.13 \\
        \bottomrule
    \end{tabular}
\end{table}

\vspace{-10pt}
\section{Conclusions}
We enable fast and efficient 2-bit LLM inference on GPUs in this paper with three novel techniques. 
We apply intra-weight mixed-precision quantization for weight matrices with 2-bit and 4-bit groups. We also introduce exclusive 16-bit sparse outliers in the 2-bit group with the minimum GPU kernel launching overhead.
We further design the asynchronous GPU kernel to accelerate LLM dequantization.
As a result, with our 2/4/16 mixed-precision quantization for each weight matrix and asynchronous dequantization during inference, our design achieves an end-to-end speedup for Llama2-7b is 1.74$\times$ over the original model, and we reduce both runtime cost and total cost by up to 2.53$\times$ and 2.29$\times$ with less GPU requirements.

\section{Acknowledgments}
This work was supported by the National Natural Science Foundation of China (No. 62104128, U21B2031), Beijing Douyin Information Service Co., Ltd.

\bibliographystyle{ACM-Reference-Format}
\bibliography{refs}


\begin{thebibliography}{45}


\ifx \showCODEN    \undefined \def \showCODEN     #1{\unskip}     \fi
\ifx \showDOI      \undefined \def \showDOI       #1{#1}\fi
\ifx \showISBNx    \undefined \def \showISBNx     #1{\unskip}     \fi
\ifx \showISBNxiii \undefined \def \showISBNxiii  #1{\unskip}     \fi
\ifx \showISSN     \undefined \def \showISSN      #1{\unskip}     \fi
\ifx \showLCCN     \undefined \def \showLCCN      #1{\unskip}     \fi
\ifx \shownote     \undefined \def \shownote      #1{#1}          \fi
\ifx \showarticletitle \undefined \def \showarticletitle #1{#1}   \fi
\ifx \showURL      \undefined \def \showURL       {\relax}        \fi
\providecommand\bibfield[2]{#2}
\providecommand\bibinfo[2]{#2}
\providecommand\natexlab[1]{#1}
\providecommand\showeprint[2][]{arXiv:#2}

\bibitem[Bisk et~al\mbox{.}(2020)]%
        {piqa}
\bibfield{author}{\bibinfo{person}{Yonatan Bisk}, \bibinfo{person}{Rowan Zellers}, {et~al\mbox{.}}} \bibinfo{year}{2020}\natexlab{}.
\newblock \showarticletitle{Piqa: Reasoning about physical commonsense in natural language}. In \bibinfo{booktitle}{\emph{AAAI}}.
\newblock


\bibitem[Camgoz et~al\mbox{.}(2020)]%
        {language_translation}
\bibfield{author}{\bibinfo{person}{Necati~Cihan Camgoz}, \bibinfo{person}{Oscar Koller}, \bibinfo{person}{Simon Hadfield}, {and} \bibinfo{person}{Richard Bowden}.} \bibinfo{year}{2020}\natexlab{}.
\newblock \showarticletitle{Sign language transformers: Joint end-to-end sign language recognition and translation}. In \bibinfo{booktitle}{\emph{Proceedings of the IEEE/CVF conference on computer vision and pattern recognition}}. \bibinfo{pages}{10023--10033}.
\newblock


\bibitem[Clark et~al\mbox{.}(2018)]%
        {arc-ec}
\bibfield{author}{\bibinfo{person}{Peter Clark}, \bibinfo{person}{Isaac Cowhey}, {et~al\mbox{.}}} \bibinfo{year}{2018}\natexlab{}.
\newblock \showarticletitle{Think you have solved question answering? try arc, the ai2 reasoning challenge}.
\newblock \bibinfo{journal}{\emph{arXiv preprint arXiv:1803.05457}} (\bibinfo{year}{2018}).
\newblock


\bibitem[Dao et~al\mbox{.}(2022)]%
        {flashattention}
\bibfield{author}{\bibinfo{person}{Tri Dao}, \bibinfo{person}{Dan Fu}, {et~al\mbox{.}}} \bibinfo{year}{2022}\natexlab{}.
\newblock \showarticletitle{Flashattention: Fast and memory-efficient exact attention with io-awareness}.
\newblock \bibinfo{journal}{\emph{NeurIPS}} (\bibinfo{year}{2022}).
\newblock


\bibitem[Dettmers et~al\mbox{.}(2023)]%
        {spqr}
\bibfield{author}{\bibinfo{person}{Tim Dettmers} {et~al\mbox{.}}} \bibinfo{year}{2023}\natexlab{}.
\newblock \showarticletitle{SpQR: A Sparse-Quantized Representation for Near-Lossless LLM Weight Compression}.
\newblock \bibinfo{journal}{\emph{arXiv preprint arXiv:2306.03078}} (\bibinfo{year}{2023}).
\newblock


\bibitem[Devlin et~al\mbox{.}(2018)]%
        {bert}
\bibfield{author}{\bibinfo{person}{Jacob Devlin}, \bibinfo{person}{Ming-Wei Chang}, \bibinfo{person}{Kenton Lee}, {and} \bibinfo{person}{Kristina Toutanova}.} \bibinfo{year}{2018}\natexlab{}.
\newblock \showarticletitle{Bert: Pre-training of deep bidirectional transformers for language understanding}.
\newblock \bibinfo{journal}{\emph{arXiv preprint arXiv:1810.04805}} (\bibinfo{year}{2018}).
\newblock


\bibitem[Du et~al\mbox{.}(2022)]%
        {nlu}
\bibfield{author}{\bibinfo{person}{Mengnan Du} {et~al\mbox{.}}} \bibinfo{year}{2022}\natexlab{}.
\newblock \showarticletitle{Shortcut learning of large language models in natural language understanding: A survey}.
\newblock \bibinfo{journal}{\emph{arXiv preprint arXiv:2208.11857}} (\bibinfo{year}{2022}).
\newblock


\bibitem[Foy(1976)]%
        {taylor}
\bibfield{author}{\bibinfo{person}{Wade~H Foy}.} \bibinfo{year}{1976}\natexlab{}.
\newblock \showarticletitle{Position-location solutions by Taylor-series estimation}.
\newblock \bibinfo{journal}{\emph{IEEE transactions on aerospace and electronic systems}} \bibinfo{number}{2} (\bibinfo{year}{1976}), \bibinfo{pages}{187--194}.
\newblock


\bibitem[Frantar and Alistarh(2023)]%
        {sparsegpt}
\bibfield{author}{\bibinfo{person}{Elias Frantar} {and} \bibinfo{person}{Dan Alistarh}.} \bibinfo{year}{2023}\natexlab{}.
\newblock \showarticletitle{SparseGPT: Massive Language Models Can Be Accurately Pruned in One-Shot}.
\newblock  (\bibinfo{year}{2023}).
\newblock


\bibitem[Frantar et~al\mbox{.}(2022)]%
        {gptq}
\bibfield{author}{\bibinfo{person}{Elias Frantar}, \bibinfo{person}{Saleh Ashkboos}, \bibinfo{person}{Torsten Hoefler}, {and} \bibinfo{person}{Dan Alistarh}.} \bibinfo{year}{2022}\natexlab{}.
\newblock \showarticletitle{Gptq: Accurate post-training quantization for generative pre-trained transformers}.
\newblock \bibinfo{journal}{\emph{arXiv preprint arXiv:2210.17323}} (\bibinfo{year}{2022}).
\newblock


\bibitem[FRED(2024)]%
        {electricity_cost}
\bibfield{author}{\bibinfo{person}{FRED}.} \bibinfo{year}{2024}\natexlab{}.
\newblock \bibinfo{title}{Average Price: Electricity per Kilowatt-Hour in U.S. City Average}.
\newblock
\newblock
\newblock
\shownote{\url{https://fred.stlouisfed.org/series/APU000072610}}.


\bibitem[Guan et~al\mbox{.}(2024)]%
        {aptq}
\bibfield{author}{\bibinfo{person}{Ziyi Guan}, \bibinfo{person}{Hantao Huang}, \bibinfo{person}{Yupeng Su}, \bibinfo{person}{Hong Huang}, \bibinfo{person}{Ngai Wong}, {and} \bibinfo{person}{Hao Yu}.} \bibinfo{year}{2024}\natexlab{}.
\newblock \showarticletitle{APTQ: Attention-aware Post-Training Mixed-Precision Quantization for Large Language Models}.
\newblock \bibinfo{journal}{\emph{arXiv preprint arXiv:2402.14866}} (\bibinfo{year}{2024}).
\newblock


\bibitem[Guo et~al\mbox{.}(2023)]%
        {greenbit}
\bibfield{author}{\bibinfo{person}{Nianhui Guo} {et~al\mbox{.}}} \bibinfo{year}{2023}\natexlab{}.
\newblock \showarticletitle{Advanced Ultra-Low Bitrate Compression Techniques for the LLaMA Family of LLMs}.
\newblock \bibinfo{journal}{\emph{https://github.com/GreenBitAI/low\_bit\_llama}} (\bibinfo{year}{2023}).
\newblock


\bibitem[Huang et~al\mbox{.}(2024)]%
        {billm}
\bibfield{author}{\bibinfo{person}{Wei Huang}, \bibinfo{person}{Yangdong Liu}, \bibinfo{person}{Haotong Qin}, \bibinfo{person}{Ying Li}, \bibinfo{person}{Shiming Zhang}, \bibinfo{person}{Xianglong Liu}, \bibinfo{person}{Michele Magno}, {and} \bibinfo{person}{Xiaojuan Qi}.} \bibinfo{year}{2024}\natexlab{}.
\newblock \showarticletitle{BiLLM: Pushing the Limit of Post-Training Quantization for LLMs}.
\newblock \bibinfo{journal}{\emph{arXiv preprint arXiv:2402.04291}} (\bibinfo{year}{2024}).
\newblock


\bibitem[HuggingFace(2024)]%
        {HuggingFace}
\bibfield{author}{\bibinfo{person}{HuggingFace}.} \bibinfo{year}{2024}\natexlab{}.
\newblock
\newblock
\newblock
\shownote{\url{https://huggingface.co/}}.


\bibitem[Keswani et~al\mbox{.}(2024)]%
        {text_summarization}
\bibfield{author}{\bibinfo{person}{Gunjan Keswani}, \bibinfo{person}{Wani Bisen}, \bibinfo{person}{Hirkani Padwad}, \bibinfo{person}{Yash Wankhedkar}, \bibinfo{person}{Sudhanshu Pandey}, {and} \bibinfo{person}{Ayushi Soni}.} \bibinfo{year}{2024}\natexlab{}.
\newblock \showarticletitle{Abstractive Long Text Summarization Using Large Language Models}.
\newblock \bibinfo{journal}{\emph{International Journal of Intelligent Systems and Applications in Engineering}} \bibinfo{volume}{12}, \bibinfo{number}{12s} (\bibinfo{year}{2024}), \bibinfo{pages}{160--168}.
\newblock


\bibitem[Kim et~al\mbox{.}(2023)]%
        {squeezellm}
\bibfield{author}{\bibinfo{person}{Sehoon Kim}, \bibinfo{person}{Coleman Hooper}, {et~al\mbox{.}}} \bibinfo{year}{2023}\natexlab{}.
\newblock \showarticletitle{SqueezeLLM: Dense-and-Sparse Quantization}.
\newblock \bibinfo{journal}{\emph{arXiv preprint arXiv:2306.07629}} (\bibinfo{year}{2023}).
\newblock


\bibitem[Li et~al\mbox{.}({[n.\,d.]})]%
        {llm-mq}
\bibfield{author}{\bibinfo{person}{Shiyao Li}, \bibinfo{person}{Xuefei Ning}, \bibinfo{person}{Ke Hong}, \bibinfo{person}{Tengxuan Liu}, \bibinfo{person}{Luning Wang}, \bibinfo{person}{Xiuhong Li}, \bibinfo{person}{Kai Zhong}, \bibinfo{person}{Guohao Dai}, \bibinfo{person}{Huazhong Yang}, {and} \bibinfo{person}{Yu Wang}.} \bibinfo{year}{[n.\,d.]}\natexlab{}.
\newblock \showarticletitle{LLM-MQ: Mixed-precision Quantization for Efficient LLM Deployment}.
\newblock  (\bibinfo{year}{[n.\,d.]}).
\newblock


\bibitem[Lin et~al\mbox{.}(2023)]%
        {awq}
\bibfield{author}{\bibinfo{person}{Ji Lin}, \bibinfo{person}{Jiaming Tang}, {et~al\mbox{.}}} \bibinfo{year}{2023}\natexlab{}.
\newblock \showarticletitle{AWQ: Activation-aware Weight Quantization for LLM Compression and Acceleration}.
\newblock \bibinfo{journal}{\emph{arXiv preprint arXiv:2306.00978}} (\bibinfo{year}{2023}).
\newblock


\bibitem[Liu et~al\mbox{.}(2024)]%
        {liu2024your}
\bibfield{author}{\bibinfo{person}{Jiawei Liu}, \bibinfo{person}{Chunqiu~Steven Xia}, \bibinfo{person}{Yuyao Wang}, {and} \bibinfo{person}{Lingming Zhang}.} \bibinfo{year}{2024}\natexlab{}.
\newblock \showarticletitle{Is your code generated by chatgpt really correct? rigorous evaluation of large language models for code generation}.
\newblock \bibinfo{journal}{\emph{Advances in Neural Information Processing Systems}}  \bibinfo{volume}{36} (\bibinfo{year}{2024}).
\newblock


\bibitem[Liu et~al\mbox{.}(2023)]%
        {llmqat}
\bibfield{author}{\bibinfo{person}{Zechun Liu}, \bibinfo{person}{Barlas Oguz}, {et~al\mbox{.}}} \bibinfo{year}{2023}\natexlab{}.
\newblock \showarticletitle{LLM-QAT: Data-Free Quantization Aware Training for Large Language Models}.
\newblock \bibinfo{journal}{\emph{arXiv preprint arXiv:2305.17888}} (\bibinfo{year}{2023}).
\newblock


\bibitem[Lund et~al\mbox{.}(2023)]%
        {lund2023chatgpt}
\bibfield{author}{\bibinfo{person}{Brady~D Lund}, \bibinfo{person}{Ting Wang}, \bibinfo{person}{Nishith~Reddy Mannuru}, \bibinfo{person}{Bing Nie}, \bibinfo{person}{Somipam Shimray}, {and} \bibinfo{person}{Ziang Wang}.} \bibinfo{year}{2023}\natexlab{}.
\newblock \showarticletitle{ChatGPT and a new academic reality: Artificial Intelligence-written research papers and the ethics of the large language models in scholarly publishing}.
\newblock \bibinfo{journal}{\emph{Journal of the Association for Information Science and Technology}} \bibinfo{volume}{74}, \bibinfo{number}{5} (\bibinfo{year}{2023}), \bibinfo{pages}{570--581}.
\newblock


\bibitem[Merity et~al\mbox{.}(2016)]%
        {wikitext}
\bibfield{author}{\bibinfo{person}{Stephen Merity}, \bibinfo{person}{Caiming Xiong}, {et~al\mbox{.}}} \bibinfo{year}{2016}\natexlab{}.
\newblock \showarticletitle{Pointer sentinel mixture models}.
\newblock \bibinfo{journal}{\emph{arXiv preprint arXiv:1609.07843}} (\bibinfo{year}{2016}).
\newblock


\bibitem[Meta(2024)]%
        {llama3}
\bibfield{author}{\bibinfo{person}{Meta}.} \bibinfo{year}{2024}\natexlab{}.
\newblock \bibinfo{title}{Build the future of AI with Meta Llama 3}.
\newblock
\newblock
\newblock
\shownote{\url{https://llama.meta.com/llama3/}}.


\bibitem[Meta(2023)]%
        {llama}
\bibfield{author}{\bibinfo{person}{AI Meta}.} \bibinfo{year}{2023}\natexlab{}.
\newblock \showarticletitle{Introducing LLaMA: A foundational, 65-billion-parameter large language model}.
\newblock \bibinfo{journal}{\emph{Meta AI}} (\bibinfo{year}{2023}).
\newblock


\bibitem[Min et~al\mbox{.}(2023)]%
        {min2023recent}
\bibfield{author}{\bibinfo{person}{Bonan Min}, \bibinfo{person}{Hayley Ross}, \bibinfo{person}{Elior Sulem}, \bibinfo{person}{Amir Pouran~Ben Veyseh}, \bibinfo{person}{Thien~Huu Nguyen}, \bibinfo{person}{Oscar Sainz}, \bibinfo{person}{Eneko Agirre}, \bibinfo{person}{Ilana Heintz}, {and} \bibinfo{person}{Dan Roth}.} \bibinfo{year}{2023}\natexlab{}.
\newblock \showarticletitle{Recent advances in natural language processing via large pre-trained language models: A survey}.
\newblock \bibinfo{journal}{\emph{Comput. Surveys}} \bibinfo{volume}{56}, \bibinfo{number}{2} (\bibinfo{year}{2023}), \bibinfo{pages}{1--40}.
\newblock


\bibitem[Nagel et~al\mbox{.}(2021)]%
        {quantwhite}
\bibfield{author}{\bibinfo{person}{Markus Nagel}, \bibinfo{person}{Marios Fournarakis}, {et~al\mbox{.}}} \bibinfo{year}{2021}\natexlab{}.
\newblock \showarticletitle{A white paper on neural network quantization}.
\newblock \bibinfo{journal}{\emph{arXiv preprint arXiv:2106.08295}} (\bibinfo{year}{2021}).
\newblock


\bibitem[Nassiri and Akhloufi(2023)]%
        {question_answer}
\bibfield{author}{\bibinfo{person}{Khalid Nassiri} {and} \bibinfo{person}{Moulay Akhloufi}.} \bibinfo{year}{2023}\natexlab{}.
\newblock \showarticletitle{Transformer models used for text-based question answering systems}.
\newblock \bibinfo{journal}{\emph{Applied Intelligence}} \bibinfo{volume}{53}, \bibinfo{number}{9} (\bibinfo{year}{2023}), \bibinfo{pages}{10602--10635}.
\newblock


\bibitem[Ni et~al\mbox{.}(2023)]%
        {nlg}
\bibfield{author}{\bibinfo{person}{Ansong Ni}, \bibinfo{person}{Srini Iyer}, {et~al\mbox{.}}} \bibinfo{year}{2023}\natexlab{}.
\newblock \showarticletitle{Lever: Learning to verify language-to-code generation with execution}. In \bibinfo{booktitle}{\emph{International Conference on Machine Learning}}.
\newblock


\bibitem[NVIDIA(2024a)]%
        {cuSPARSE}
\bibfield{author}{\bibinfo{person}{NVIDIA}.} \bibinfo{year}{2024}\natexlab{a}.
\newblock \bibinfo{title}{NVIDIA CUDA Sparse Matrix Library}.
\newblock
\newblock
\newblock
\shownote{\url{https://docs.nvidia.com/cuda/cusparse/}}.


\bibitem[NVIDIA(2024b)]%
        {nvml}
\bibfield{author}{\bibinfo{person}{NVIDIA}.} \bibinfo{year}{2024}\natexlab{b}.
\newblock \bibinfo{title}{NVIDIA Management Library (NVML) | NVIDIA Developer}.
\newblock
\newblock
\newblock
\shownote{\url{https://developer.nvidia.com/nvidia-management-library-nvml}}.


\bibitem[Patel and Ahmad(2024)]%
        {openai_cost}
\bibfield{author}{\bibinfo{person}{Dylan Patel} {and} \bibinfo{person}{Afzal Ahmad}.} \bibinfo{year}{2024}\natexlab{}.
\newblock \bibinfo{title}{The Inference Cost Of Search Disruption - Large Language Model Cost Analysis}.
\newblock
\newblock
\newblock
\shownote{\url{https://www.semianalysis.com/p/the-inference-cost-of-search-disruption}}.


\bibitem[Peng et~al\mbox{.}(2023)]%
        {gpt4}
\bibfield{author}{\bibinfo{person}{Baolin Peng}, \bibinfo{person}{Chunyuan Li}, {et~al\mbox{.}}} \bibinfo{year}{2023}\natexlab{}.
\newblock \showarticletitle{Instruction tuning with gpt-4}.
\newblock \bibinfo{journal}{\emph{arXiv preprint arXiv:2304.03277}} (\bibinfo{year}{2023}).
\newblock


\bibitem[Sakaguchi et~al\mbox{.}(2021)]%
        {winogrande}
\bibfield{author}{\bibinfo{person}{Keisuke Sakaguchi}, \bibinfo{person}{Ronan~Le Bras}, {et~al\mbox{.}}} \bibinfo{year}{2021}\natexlab{}.
\newblock \showarticletitle{Winogrande: An adversarial winograd schema challenge at scale}.
\newblock \bibinfo{journal}{\emph{Commun. ACM}} (\bibinfo{year}{2021}).
\newblock


\bibitem[Schaeffer et~al\mbox{.}(2024)]%
        {schaeffer2024emergent}
\bibfield{author}{\bibinfo{person}{Rylan Schaeffer}, \bibinfo{person}{Brando Miranda}, {and} \bibinfo{person}{Sanmi Koyejo}.} \bibinfo{year}{2024}\natexlab{}.
\newblock \showarticletitle{Are emergent abilities of large language models a mirage?}
\newblock \bibinfo{journal}{\emph{Advances in Neural Information Processing Systems}}  \bibinfo{volume}{36} (\bibinfo{year}{2024}).
\newblock


\bibitem[Shao et~al\mbox{.}(2023)]%
        {omniquant}
\bibfield{author}{\bibinfo{person}{Wenqi Shao}, \bibinfo{person}{Mengzhao Chen}, \bibinfo{person}{Zhaoyang Zhang}, \bibinfo{person}{Peng Xu}, \bibinfo{person}{Lirui Zhao}, \bibinfo{person}{Zhiqian Li}, \bibinfo{person}{Kaipeng Zhang}, \bibinfo{person}{Peng Gao}, \bibinfo{person}{Yu Qiao}, {and} \bibinfo{person}{Ping Luo}.} \bibinfo{year}{2023}\natexlab{}.
\newblock \showarticletitle{Omniquant: Omnidirectionally calibrated quantization for large language models}.
\newblock \bibinfo{journal}{\emph{arXiv preprint arXiv:2308.13137}} (\bibinfo{year}{2023}).
\newblock


\bibitem[Shen et~al\mbox{.}(2020)]%
        {qbert}
\bibfield{author}{\bibinfo{person}{Sheng Shen}, \bibinfo{person}{Zhen Dong}, \bibinfo{person}{Jiayu Ye}, \bibinfo{person}{Linjian Ma}, \bibinfo{person}{Zhewei Yao}, \bibinfo{person}{Amir Gholami}, \bibinfo{person}{Michael~W Mahoney}, {and} \bibinfo{person}{Kurt Keutzer}.} \bibinfo{year}{2020}\natexlab{}.
\newblock \showarticletitle{Q-bert: Hessian based ultra low precision quantization of bert}. In \bibinfo{booktitle}{\emph{Proceedings of the AAAI Conference on Artificial Intelligence}}, Vol.~\bibinfo{volume}{34}. \bibinfo{pages}{8815--8821}.
\newblock


\bibitem[Touvron et~al\mbox{.}(2023)]%
        {llama2}
\bibfield{author}{\bibinfo{person}{Hugo Touvron}, \bibinfo{person}{Louis Martin}, {et~al\mbox{.}}} \bibinfo{year}{2023}\natexlab{}.
\newblock \showarticletitle{Llama 2: Open foundation and fine-tuned chat models}.
\newblock \bibinfo{journal}{\emph{arXiv preprint arXiv:2307.09288}} (\bibinfo{year}{2023}).
\newblock


\bibitem[Vaswani et~al\mbox{.}(2017)]%
        {transformer}
\bibfield{author}{\bibinfo{person}{Ashish Vaswani}, \bibinfo{person}{Noam Shazeer}, \bibinfo{person}{Niki Parmar}, \bibinfo{person}{Jakob Uszkoreit}, \bibinfo{person}{Llion Jones}, \bibinfo{person}{Aidan~N Gomez}, \bibinfo{person}{{\L}ukasz Kaiser}, {and} \bibinfo{person}{Illia Polosukhin}.} \bibinfo{year}{2017}\natexlab{}.
\newblock \showarticletitle{Attention is all you need}.
\newblock \bibinfo{journal}{\emph{Advances in neural information processing systems}}  \bibinfo{volume}{30} (\bibinfo{year}{2017}).
\newblock


\bibitem[Wang et~al\mbox{.}(2018)]%
        {glue}
\bibfield{author}{\bibinfo{person}{Alex Wang}, \bibinfo{person}{Amanpreet Singh}, \bibinfo{person}{Julian Michael}, \bibinfo{person}{Felix Hill}, \bibinfo{person}{Omer Levy}, {and} \bibinfo{person}{Samuel~R Bowman}.} \bibinfo{year}{2018}\natexlab{}.
\newblock \showarticletitle{GLUE: A multi-task benchmark and analysis platform for natural language understanding}.
\newblock \bibinfo{journal}{\emph{arXiv preprint arXiv:1804.07461}} (\bibinfo{year}{2018}).
\newblock


\bibitem[Wei et~al\mbox{.}(2022)]%
        {emergent}
\bibfield{author}{\bibinfo{person}{Jason Wei}, \bibinfo{person}{Yi Tay}, \bibinfo{person}{Rishi Bommasani}, \bibinfo{person}{Colin Raffel}, \bibinfo{person}{Barret Zoph}, \bibinfo{person}{Sebastian Borgeaud}, \bibinfo{person}{Dani Yogatama}, \bibinfo{person}{Maarten Bosma}, \bibinfo{person}{Denny Zhou}, \bibinfo{person}{Donald Metzler}, {et~al\mbox{.}}} \bibinfo{year}{2022}\natexlab{}.
\newblock \showarticletitle{Emergent abilities of large language models}.
\newblock \bibinfo{journal}{\emph{arXiv preprint arXiv:2206.07682}} (\bibinfo{year}{2022}).
\newblock


\bibitem[Xiao et~al\mbox{.}(2023)]%
        {smoothquant}
\bibfield{author}{\bibinfo{person}{Guangxuan Xiao}, \bibinfo{person}{Ji Lin}, {et~al\mbox{.}}} \bibinfo{year}{2023}\natexlab{}.
\newblock \showarticletitle{Smoothquant: Accurate and efficient post-training quantization for large language models}. In \bibinfo{booktitle}{\emph{ICML}}.
\newblock


\bibitem[Zadeh et~al\mbox{.}(2020)]%
        {gobo}
\bibfield{author}{\bibinfo{person}{Ali~Hadi Zadeh}, \bibinfo{person}{Isak Edo}, \bibinfo{person}{Omar~Mohamed Awad}, {and} \bibinfo{person}{Andreas Moshovos}.} \bibinfo{year}{2020}\natexlab{}.
\newblock \showarticletitle{Gobo: Quantizing attention-based nlp models for low latency and energy efficient inference}. In \bibinfo{booktitle}{\emph{2020 53rd Annual IEEE/ACM International Symposium on Microarchitecture (MICRO)}}. IEEE, \bibinfo{pages}{811--824}.
\newblock


\bibitem[Zellers et~al\mbox{.}(2019)]%
        {hellaswag}
\bibfield{author}{\bibinfo{person}{Rowan Zellers}, \bibinfo{person}{Ari Holtzman}, {et~al\mbox{.}}} \bibinfo{year}{2019}\natexlab{}.
\newblock \showarticletitle{Hellaswag: Can a machine really finish your sentence?}
\newblock \bibinfo{journal}{\emph{arXiv preprint arXiv:1905.07830}} (\bibinfo{year}{2019}).
\newblock


\bibitem[Zeng et~al\mbox{.}(2022)]%
        {chatglm3}
\bibfield{author}{\bibinfo{person}{Aohan Zeng}, \bibinfo{person}{Xiao Liu}, {et~al\mbox{.}}} \bibinfo{year}{2022}\natexlab{}.
\newblock \showarticletitle{Glm-130b: An open bilingual pre-trained model}.
\newblock \bibinfo{journal}{\emph{arXiv preprint arXiv:2210.02414}} (\bibinfo{year}{2022}).
\newblock


\end{thebibliography}

\end{document}